\documentclass[lettersize,journal]{IEEEtran}
\usepackage{amsmath,amsfonts}
\usepackage{algorithmic}
\usepackage{algorithm}
\usepackage{array}
\usepackage[caption=false,font=scriptsize,labelfont=rm,textfont=rm]{subfig}
\usepackage{textcomp}
\usepackage{stfloats}
\usepackage{url}
\usepackage{verbatim}
\usepackage{graphicx}
\usepackage{cite}

\usepackage{amsthm}
\usepackage{anyfontsize}
\usepackage{float}            
\usepackage{mathrsfs}         
\usepackage{stfloats}
\usepackage{arydshln}
\usepackage{bm}               
\usepackage{threeparttable}   
\usepackage{makecell}

\usepackage{booktabs}
\usepackage{multirow}
\usepackage{wrapfig}
\usepackage{balance}
\usepackage[table]{xcolor}
\usepackage{colortbl}

\usepackage[colorlinks,bookmarksopen,bookmarksnumbered,citecolor=blue, linkcolor=blue, urlcolor=blue]{hyperref}

\newcommand{\down}[1]{{\color{gray}($\downarrow$#1)}}
\newcommand{\up}[1]{{\color{blue}($\uparrow$#1)}}

\newcommand{\tokenlearned}[1]{#1~(Learned)}
\newcommand{\tokenfixed}[1]{#1~(Fixed)}

\newcommand{\ie}[0]{\emph{i.e.}}
\newcommand{\eg}[0]{\emph{e.g.}}
\newcommand{\sysname}[0]{\textcolor{black}{PRANCE}}

\newcommand{\R}{\ensuremath{\mathbb{R}}}

\begin{document}

\title{\sysname: Joint Token-Optimization and Structural Channel-Pruning for Adaptive ViT Inference}

\author{Ye Li$^{*}$,
        Chen Tang$^{*}$,
        Yuan Meng$^{\dagger}$,
        Jiajun Fan, 
        Zenghao Chai, 
        Xinzhu Ma, 
        Zhi Wang$^{\dagger}$,~\IEEEmembership{Senior Member,~IEEE}, \\
        Wenwu Zhu$^{\dagger}$,~\IEEEmembership{Fellow,~IEEE}
\IEEEcompsocitemizethanks{
\IEEEcompsocthanksitem Ye Li, Chen Tang, Yuan Meng, Zenghao Chai, Zhi Wang, Wenwu Zhu are with Tsinghua University, China. 
Ye Li and Zhi Wang are also with Tsinghua-Berkeley Shenzhen Institute, Tsinghua Shenzhen International Graduate School, Tsinghua University. 
Xinzhu Ma is with MMLab, The Chinese University of Hong Kong, China. 
\IEEEcompsocthanksitem $^{*}$Equal contribution. ~~\textsuperscript{\dag}Corresponding author. 
}
}



\maketitle

\begin{abstract}
The troublesome model size and quadratic computational complexity associated with token quantity pose significant deployment challenges for Vision Transformers (ViTs) in practical applications. 
Despite recent advancements in model pruning and token reduction techniques speed up the inference speed of ViTs, these approaches either adopt a fixed sparsity ratio or overlook the meaningful interplay between architectural optimization and token selection. 
Consequently, this \emph{static} and \emph{single-domain} compression often leads to pronounced accuracy degradation under aggressive compression rates, as they fail to fully explore redundancies across these two orthogonal domains.
To address these issues, we introduce \sysname, a framework that jointly optimizes the activated channels and reduces tokens, based on the characteristics of inputs. 
Specifically, \sysname~ leverages adaptive token optimization strategies for a certain computational budget, aiming to accelerate ViTs' inference from a unified data and architectural perspective.
However, the joint framework poses challenges to both architectural and decision-making aspects. 
Firstly, while ViTs inherently support variable-token inference, they do not facilitate dynamic computations for variable channels.
To overcome this limitation, we propose a meta-network using weight-sharing techniques to support arbitrary channels of the Multi-head Self-Attention and Multi-layer Perceptron layers, serving as a foundational model for architectural decision-making.
Second, simultaneously optimizing the structure of the meta-network and input data constitutes a combinatorial optimization problem with an extremely large decision space, reaching up to around $10^{14}$, making supervised learning infeasible.
To this end, we design a lightweight selector employing Proximal Policy Optimization for efficient decision-making. Furthermore, we introduce a novel "Result-to-Go" training mechanism that models ViTs' inference process as a Markov decision process, significantly reducing action space and mitigating delayed-reward issues during training.
Extensive experiments demonstrate the effectiveness of \sysname~ in reducing FLOPs by approximately 50\%, retaining only about 10\% of tokens while achieving lossless Top-1 accuracy. Additionally, our framework is shown to be compatible with various token optimization techniques such as pruning, merging, and sequential pruning-merging strategies.
The code is available at \href{https://github.com/ChildTang/PRANCE}{https://github.com/ChildTang/PRANCE}.

\end{abstract}


\section{Introduction}
\IEEEPARstart{V}{ision} Transformers (ViTs) \cite{han2022survey} have emerged as cutting-edge architectures across various fields of machine learning, including classification \cite{vitclass}, detection \cite{vitdetect,vit-detection-2}, segmentation \cite{vitseg,seg-2}, multi-modal modeling \cite{vitmm,seg-2,vlt-2}, {\it etc.}
Multi-Head Self-Attention (MHSA), the core of the Transformers, empowers global representation modeling by dynamically weighting each token within the input sequence. 
However, the quadratic increased complexity of MHSA with respect to the number of tokens, coupled with the model size, significantly exacerbates deployment challenges.

\begin{figure}[t]
\centering
\includegraphics[scale=0.44]{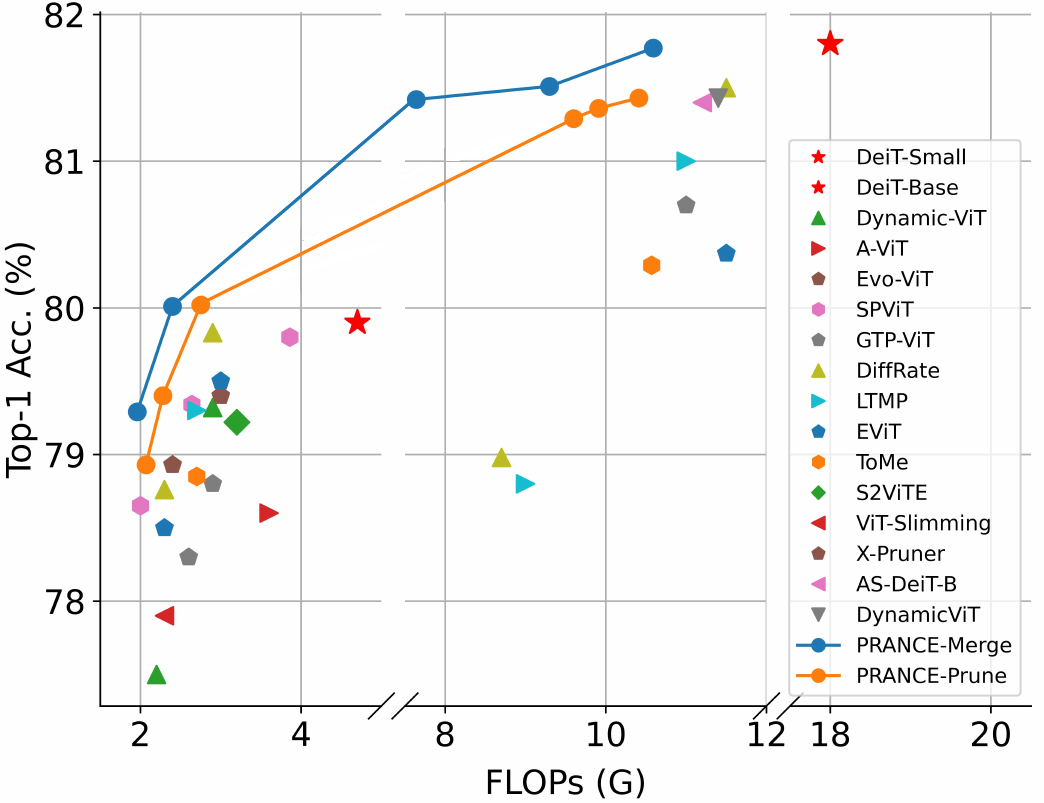}
\caption{\textbf{Comparison of \sysname~with SOTA methods.} \sysname~ achieves both higher Top-1 accuracy and lower complexity (FLOPs) in ImageNet.}
\vspace{-8pt}
\end{figure}

The computational overheads of ViTs are primarily concentrated in  \textit{(i)} the number of embedding dimensions $C$ and \textit{(ii)} the quadratic complexity with the number of tokens $N$. 
Therefore, there have been two concurrent research directions that aim to improve the efficiency of ViTs including \emph{(i)} structure compression and \emph{(ii)} token optimization. 
As one of the most direct ways, the former one leverages the conventional deep model compression techniques, \ie, model pruning \cite{vit-pruning,vit_wd_pruning}, weight quantization \cite{ptq-vit,tang2022mixed,tang2024retraining}, and lightweight model design \cite{Autoformer,Elasticvit,yu2020bignas}, to remove the redundant components of ViTs. 
For example, NViT \cite{nvit} adopts structural pruning in the fields of CNNs compression to prune the channels with Hessian information. 
ElasticViT \cite{Elasticvit} and AutoFormer \cite{Autoformer} automate the design process of ViTs with the help of two-stage neural architecture search \cite{yu2020bignas,wang2021attentivenas}. 
On the other hand, token optimization methods work on directly manipulating the number of tokens with a \emph{predefined} token keep ratio, which is a kind of Transformer-specific technique in contrast to model compression due to the support of variable token length in MHSA. 
Specifically, token optimization methods can be divided into pruning-based methods and merging-based methods. 
Pruning-based methods \cite{dynamicvit,liang2022evit} remove the uninformative tokens progressively during inference according to the calculated importance score. 
For example, 
DynamicViT \cite{dynamicvit} incorporates a lightweight Multi-layer Perceptron (MLP) layer \cite{haykin1994neural,nipsw23} for evaluating and pruning token values. 
Merging-based methods \cite{tome,bolya2023token} reduce token quantity by progressively merging tokens with high similarities during the inference process, {\it e.g.,} ToMe \cite{tome} measures token correlations based on the cosine similarity of their K matrices in the attention mechanism and merges them by calculating mean values.
Based on the above methods, several advanced evaluation mechanisms and fusion methods \cite{diffrate,long2023beyond,wang2023zero} have been proposed to achieve higher compression rates while maintaining accuracy.

\begin{figure*}[t] 
    \centering
    \includegraphics[width=1.0\textwidth]{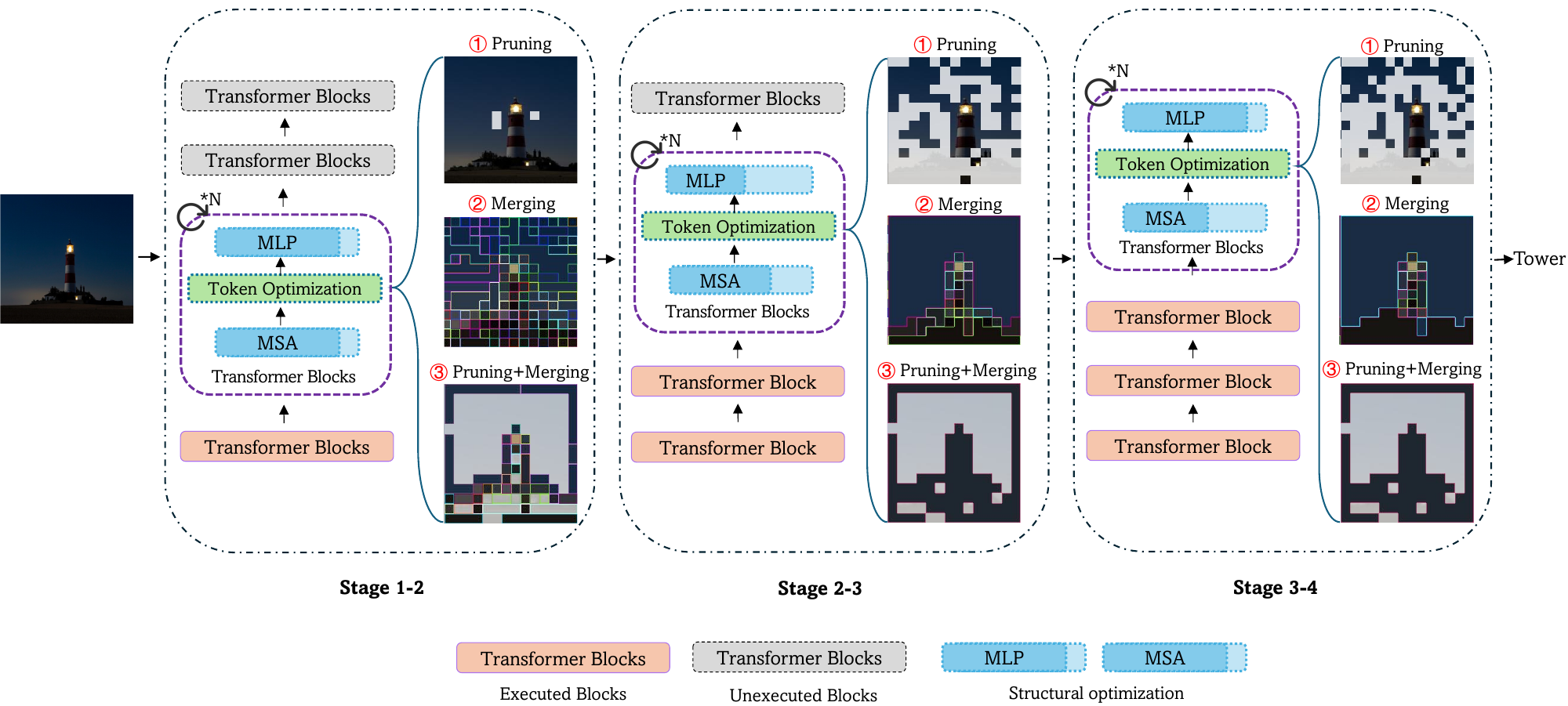}
    \caption{
    \textbf{Illustration of the inference process of \sysname.} \sysname~is a lightweight framework for ViTs that jointly optimizes model structure and data. First of all, the framework divides the ViT model into four groups according to the inference sequence, each containing multiple ViT blocks. During inference, the selector utilizes the features of each group step by step to decide the model channel dimensions and token numbers for them, aiming to minimize FLOPs while ensuring accuracy. Moreover, \sysname~supports three main token optimization methods: pruning, merging, and pruning-then-merging. 
    }
    \label{data_flow}
\vspace{-8pt}
\end{figure*}

While significant reductions in complexity (\eg, FLOPs, latency, model size, {\it etc.}) have been achieved, the above methods still have some limitations. 
From the perspective of structural reduction, they typically search for several independent low-complexity models for various downstream tasks by resorting to \emph{data-agnostic} methods, therefore remaining significant redundancies in the model structures when applied to different data samples. 
In other words, they ignore that samples with varying recognition difficulties often carry different amounts of useful information. For example, consider two images of the same size: one is an apple against a white wall, and the other is a vibrant cityscape at night. 
Clearly, the second image contains significantly more information than the first, and thus the uninformative input should be applied more simplified architecture for efficiency. 
That is to say, the number of channels utilized by ViTs should dynamically adjust when processing images of varying complexity. 
From the perspective of data optimization, as the Transformer natively supports variable input lengths, recent studies tend to progressively evict the uninformative tokens during the inference process for the inputs, and the eviction tokens will be directly removed \cite{dynamicvit} or merged into other tokens \cite{tome}. 
However, token optimization typically faces significant performance degradation in the high compression ratio, this makes recent studies employ rather sophisticated token matching and reduction techniques (e.g., NAS-based search \cite{diffrate}) to retain performance but inevitably compromise the runtime efficiency and implementation simplicity. 
What's more, whether optimizing the model architecture or the data, the essence lies in eliminating redundant data and preserving effective data to achieve the optimal trade-off between model accuracy and FLOPs. 
Therefore, they are not entirely orthogonal and cannot be simply combined. Based on these works, we would like to ask a question: 

\noindent \textbf{\textit{Whether it is possible to adaptively optimize both the model architecture and tokens for each sample simultaneously to achieve the optimal accuracy-FLOPs trade-off?}}

To answer this question, two key issues need to be resolved:  
\emph{(i)} although ViTs inherently support a variable number of tokens, they do not support a variable dimension of channels; and \emph{(ii)} the optimization space created by combined optimization is excessively large, making it difficult to find the optimal solution. 
It is non-trivial to search for an optimal compression ratio for both architectural and token-level optimization by naively combining the existing methods in these two fields, as the resulting decision space can reach up to $6 \times {10^{12}} $. 
In this paper, we propose the \sysname~framework to address the above challenges. By optimizing both the channels and tokens from the perspectives of model architecture and data, we aim to minimize FLOPs while ensuring accuracy across different samples, the overall inference process is shown in Fig.~\ref{data_flow}.

\sysname~ starts with a meta-network training process to support arbitrary channels of the MHSA and MLP layer. 
We adopt the weight-sharing technique to allow the smaller channels to be a subset of the large channels~\cite{Elasticvit,Autoformer}, and thus the weights of different channel candidates can be coupled and learned together. 
To simulate the architectural decision made by the architectural selector we introduced after, we perform random sampling to select different architectures at each step of meta-network training, and the resulting model could receive variable channels after convergence. 
It is noteworthy that the training of meta-network is only performed once and uses all tokens by default (as in conventional ViTs training), which aims to improve the training stability. 
Then, to solve the combinatorial optimization problem in the mixed-decision space, where the decision values are divided into channel dimension selection and token optimization ratio decisions, we consider leveraging Proximal Policy Optimization (PPO) to conduct efficiency learning. 
By modeling the decision-making process of architecture and tokens as a Markov process, along with a newly designed ``Result-to-Go'' mechanism, we achieve accurate decision estimation for each action taken by the selector. 
Moreover, we further experiment with three primary token optimization strategies: pruning, merging, and pruning-then-merging, and show that our frameworks can seamlessly be compatible with them. 
Surprisingly, we have observed even the simplest token optimization strategy can surpass previous advanced methods, which further demonstrates the effectiveness of the proposed framework. 
For example, \sysname~achieved Top-1 accuracies of 72.38\%, 79.98\%, and 81.77\% on ViT-tiny, ViT-small, and ViT-base models, respectively, while requiring only 0.87 GFLOPs, 2.38 GFLOPs, and 10.59 GFLOPs and saving up to 53\% FLOPs. 
We build up a new Pareto front of the Vision Transformer compressions, which sheds light on the importance of joint optimization for both architectural and data aspects. 

To summarize, our contributions are as follows: 
\begin{itemize}
\item [(1)] We propose a high-accuracy, low-FLOPs framework, \sysname, for ViTs compression, which allows sample-wise joint optimization of token numbers $N$ and channel dimensions $C$ during inference. The proposed \sysname~optimizes the model from both the structural and data perspectives. For samples with different complexities, \sysname~dynamically selects the channels and tokens with more information during inference to achieve the optimal efficiency-accuracy trade-off. 
\item [(2)] We construct a meta-network with variable channels. Specifically, we train a high-performing meta-network using weight-sharing techniques under multiple selectable MHSA channels and MLP expansion ratios, providing \sysname~with a foundational model capable of dynamically adjusting channel dimensions and token numbers. 
\item [(3)] We propose a lightweight PPO-based selector for ViTs and introducing the new training mechanism ``Result-to-Go", which significantly reduces the selection space of combinatorial optimization problems. The proposed selector can progressively optimize the data and model structures while maintaining the compact action space dimensions. 
\item [(4)] We conduct extensive experiments and demonstrate that \sysname~exhibits excellent performance. It achieves higher Top-1 accuracy with lower FLOPs across Tiny, Small, and Base scales, surpassing various state-of-the-art methods. Moreover, it also supports the joint optimization of model structures with three data optimization methods: pruning, merging, and pruning-then-merging. 
\end{itemize}

\section{Related Work}
\noindent \textbf{Vision Transformers (ViTs) Compression.} 
To deploy ViTs on resource-limited devices, recent advances lean upon the conventional model compression techniques to lessen the \emph{architectural redundancies} of these over-parameterized models, including model pruning~\cite{nvit,vit-pruning,vit_wd_pruning,vit-slimming}, quantization~\cite{ptq-vit,fq-vit,tang2022arbitrary,tang2022mixed}, and lightweight architecture design~\cite{Autoformer,Elasticvit,Efficientvit}. 
They explore effective solutions for lightweight ViTs from three aspects: reducing the number of model channels, lowering the precision of model storage, and finding more optimal lightweight structures. 
Specifically, NViT~\cite{nvit} constructs the final model by establishing a global importance score ranking, observing the trend of dimension changes in the pruned network structure, and reallocating parameters. 
FQ-ViT~\cite{fq-vit} adaptively quantizes all structures of the ViTs model while maintaining results similar to those of the full precision model. 
ElasticViT \cite{Elasticvit} improves the sampling strategies form the Autoformer \cite{Autoformer} for the supernet training, designing a high-precision, lightweight ViTs model that supports a wide range of mobile devices with varying computational power. 
The above methods explore the possibilities of architectural optimization for ViTs in various domains and achieve excellent results at the task level. 
However, it is evident that the requirements for model parameters and quantization bit-width should differ when processing simple and complex images with ViTs. 
Additionally, these methods overlook optimization in the data dimension: ViTs can achieve accurate results by focusing only on the important parts of different samples, while the unimportant parts lead to redundant computations.
Therefore, there is tremendous potential for optimization by jointly optimizing the structure dimension and the data dimension at the sample level.

\noindent \textbf{Token Optimization for Vision Transformers.} 
From the perspective of data optimization, reducing the less informative parts of the data can effectively lower computational complexity while maintaining model performance. 
Unlike CNN-based models, which extract image features through convolution, transformer-based models encode data into tokens and get the semantics of them through attention mechanism. 
Consequently, their computational complexity grows quadratically with the number of tokens.
Therefore, extensive efforts have been made to reduce the token number, and these efforts can be divided into two primary paradigms: token pruning \cite{token_pruning_initial,A-vit,dynamicvit,Evo_vit} and token merging \cite{tome,bolya2023token,diffrate,multi_scale_token,token_pooling,merge_token_vit}.
E-ViT \cite{liang2022evit} believes that the importance of tokens is reflected in the \texttt{<CLS>} token. Therefore, it sorts the tokens based on the value of the \texttt{<CLS>} token, retaining the important tokens based on the manually defined rate and merging the less important ones into a single token.
DynamicViT \cite{dynamicvit} removes the tokens by inserting an additional MLP layer in adjacent ViTs blocks to learn the pruning decision, and achieves sample-wise token pruning. 
On the other hands, the merging-based approach posits that directly discarding tokens results in information loss. By merging similar tokens, we can reduce the number of tokens while preserving more information.
ToMe \cite{tome,bolya2023token} utilize the Bipartite Soft Matching to calculate the distance between tokens and then merge the similar tokens. 
Besides, recent methods typically require more complex token assessment and matching techniques. 
For example, Diffrate \cite{diffrate} utilizes pruning followed by merging to further improve compression rates while maintaining good accuracy.
BAT~\cite{long2023beyond} needs a matching-then-clustering strategy to identify the importance and diversity of tokens, and Zero-TPrune~\cite{wang2023zero} develops a graph-based matching to calculate the similarity of tokens.  
However, the above methods only consider data optimization, ignoring the model structural redundancy and the coupling between data and structure. Therefore, there remains significant optimization space for ViTs.

In this paper, we have demonstrated that by using only basic pruning, merging, and combinatorial methods, complemented by joint architectural optimization, we can also achieve superior performance-efficiency trade-offs compared to previous methods. Our approach further eschews the intricate token-to-token matching mechanisms required by merging techniques, which could slow down the inference on real hardware.

\section{Method}

\begin{figure*}[t]
    \centering
    \includegraphics[width=1.0\textwidth]{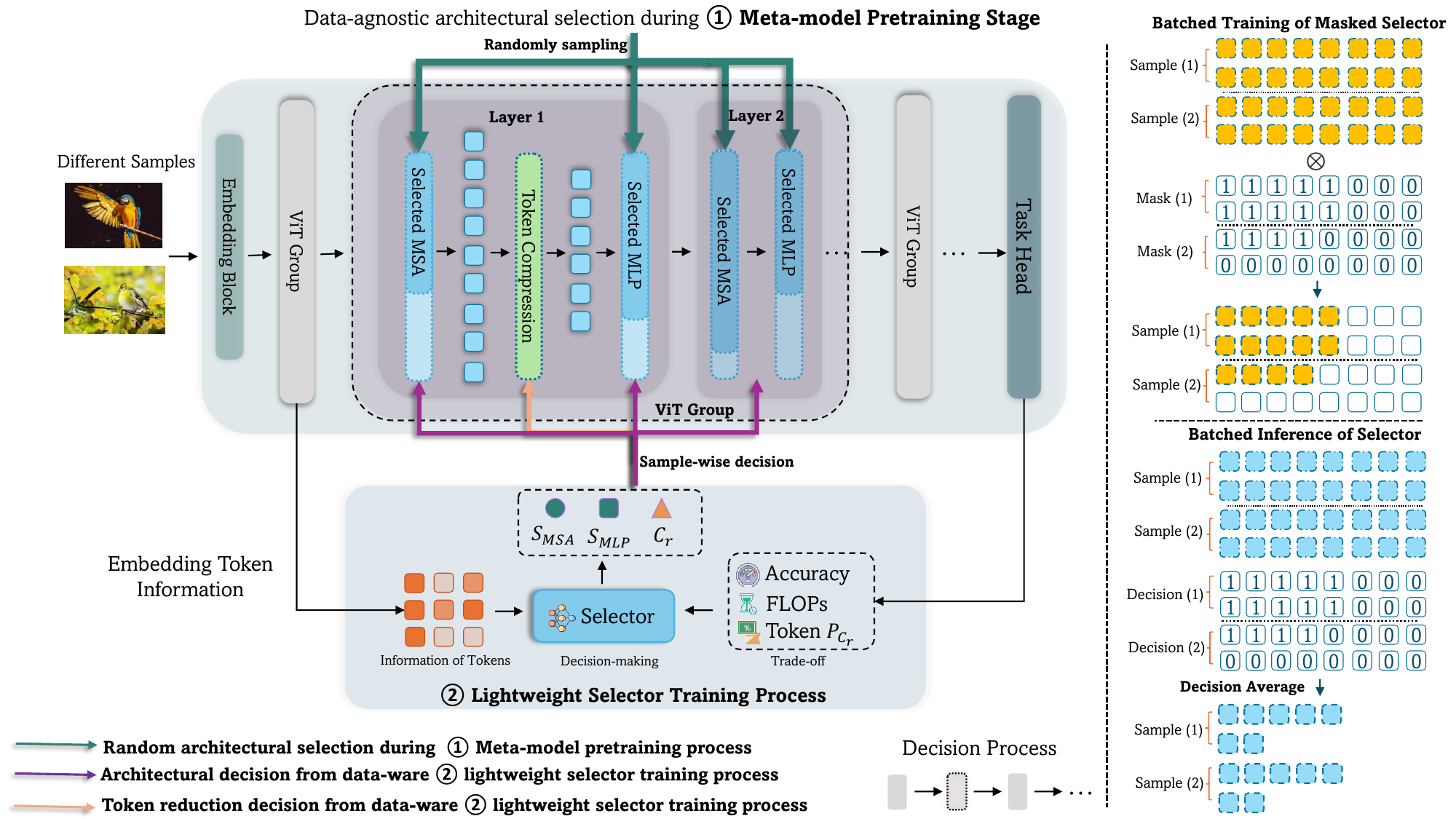}
    \caption{
    \textbf{The framework of \sysname.} 
    \emph{\textbf{Left:}} The training of \sysname~consists of two stages: 
    (1) Meta-model Pretraining. The meta-network is trained using the weight-sharing mechanism, where the smaller channels are subsets of the large channels, to support the variable channels. 
    To simulate the variable channel decisions, a configuration is randomly selected for the MHSA layer and MLP layer in each training step. In this stage, we do not perform token optimization. 
    (2) Sample-wise architecture-data joint optimization. After convergence of the meta-network, we freeze the meta-network and train the PPO selector using the ``Result-to-Go'' mechanism. In this stage, the PPO selector will jointly make the decisions for channel reduction of the MHSA layer and MLP layer, along with the decision of token reduction. 
    \emph{\textbf{Right:}} We adopt a sample-wise masking mechanism for supporting batched training of the selector, where the decisions are generated in the form of 0-1 mask and applied on the corresponding inputs (e.g., tokens, channels) using Hadamard product accordingly, to ensure dimensional consistency. 
    During inference, the sample-wise mask can be replaced by averaging the decisions of each batch. 
    }
    \label{framework}
\end{figure*}

The overall framework of \sysname~is shown in Fig.~\ref{framework}, which involves two steps: 
firstly, pretraining a meta-network of ViTs with variable channels through simulated channel selection decisions, 
secondly, segmenting every three blocks of the ViTs into distinct groups, integrating a PPO-based lightweight selector between groups for conducting sample-wise architectural decisions and token selections, and training the selector through Reinforcement Learning (RL). After that, we also conduct fine-tuning of the ViTs backbone to align the representation of ViTs and the selector for optimal performance. 

\subsection{Preliminary of Computation Complexity}
The computation of the Transformer-based models mainly consists of two parts: MHSA and Feed-Forward Network (FFN). Suppose the input dimension is $(N, C)$, where $N$ is the token number and $C$ is the embedding dimension of the token. Then the computational complexity of MHSA is $\mathcal{O}(4N{C^2} + 2{N^2}C)$, the computational complexity of FFN is $\mathcal{O}(8N{C^2})$, and the total computational complexity is $\mathcal{O}(12N{C^2} + 2{N^2}C)$. Reducing the computational complexity of such models requires attention to two aspects: from the perspective of model architecture, we can optimize the token length $C$, and from the perspective of data, we can optimize the number of tokens $N$. Additionally, considering that the effective number of tokens and the required model architecture may vary for different samples, performing sample-wise optimization on these two aspects can ideally yield optimal results.

\subsection{Channel-elastic Meta-network Training}
To support channel selection on different-sized ViTs models (\eg, ViT-tiny, ViT-small, ViT-base, etc), we train several meta ViTs with variable channel and MLP ratio by making the ViTs perceive the architectural changes. 
Specifically, we use a set of pre-defined embedding dimensions and MLP ratios of the model, as shown in Tab.~\ref{tab:supernet_setting}. 
To support the variable channel, we enable the MHSA layers can be assigned a specific embedding dimension $\phi$: 
\begin{equation}
    \texttt{MHSA}(\mathbf{x}; \phi) =  \texttt{Softmax} \left( \frac{{\mathbf{Q}_{\phi}(\mathbf{K}_{\phi})^\mathsf{T}}}{\sqrt{\phi}} \right)\mathbf{V}_{\phi}, 
\end{equation} 
where $\mathbf{Q}_{\phi} \in \R^{(N+1) \times \phi}$, $\mathbf{K}_{\phi} \in \R^{(N+1) \times \phi}$ and $\mathbf{V}_{\phi} \in \R^{(N+1) \times \phi}$ are projected matrices with a given embedding dimension of ${\phi}$ and the input $\mathbf{x} \in \R^{(N+1) \times C_{in}}$: 
\begin{equation} 
\footnotesize
    \mathbf{Q}_{\phi} = \mathbf{x}(\mathbf{W}^{q}_{\phi})^{\mathsf{T}} \qquad 
    \mathbf{K}_{\phi} = \mathbf{x}(\mathbf{W}^{k}_{\phi})^{\mathsf{T}} \qquad 
    \mathbf{V}_{\phi} = \mathbf{x}(\mathbf{W}^{v}_{\phi})^{\mathsf{T}}, 
\end{equation}
where the projection weights $\mathbf{W}^{q}_{\phi} \in \R^{\phi \times {C_{in}}}$, $\mathbf{W}^{k}_{\phi} \in \R^{\phi \times C_{in}}$, $\mathbf{W}^{v}_{\phi} \in \R^{\phi \times C_{in}}$ are sliced from their full weights using first $\phi$ channels \cite{Elasticvit,yu2020bignas}: 
\begin{equation}
\footnotesize
    \mathbf{W}^{q}_{\phi} = \mathbf{W}^{q} \left[ :\phi, : C_{in} \right] \quad 
    \mathbf{W}^{k}_{\phi} = \mathbf{W}^{k} \left[ :\phi, : C_{in} \right] \quad 
    \mathbf{W}^{v}_{\phi} = \mathbf{W}^{v} \left[ :\phi, : C_{in} \right]. \quad
    \label{_eq:sliced_projection}
\end{equation}
In the meta-network training stage, to simulate channel selection decisions, an embedding dimension $\phi = E$ is randomly sampled at each training step and adopted to the MHSA layer. 

Similarly, the channel-variable MLP consists of two linear projections and one non-linear activation function, where the first linear projection is used to expand the channels by a factor of $\gamma$ \cite{vitclass,vaswani2017attention}, and the second linear projection is used to restore the channels, which can be represented as: 
\begin{equation}
    \texttt{MLP}(\mathbf{x}; \gamma) =   
 {\texttt{GeLU} \left( \mathbf{x} \left(\mathbf{W}^{\tt UP}_{\gamma} \right)^{\mathsf{T}} \right) } {\left(\mathbf{W}^{\tt DOWN}_{\gamma} \right)}^{\mathsf{T}}. 
\end{equation} 
Similar to Eq. \ref{_eq:sliced_projection}, we slice the first $\gamma$ times input channels from the full weights $\mathbf{W}^{\tt UP}$ and $\mathbf{W}^{\tt DOWN}$ to obtain $\mathbf{W}^{\tt UP}_{\gamma}$ and $\mathbf{W}^{\tt DOWN}_{\gamma}$. 

The trained meta-network provides the basis our method to adjust the embedding dimension. 
In the following parts, we will describe how to optimize token number and token length through PPO selector and achieve the best trade-off between sample-wise accuracy and computational complexity.

\subsection{Sample-wise Joint Optimization with Lightweight PPO Agent}
After obtaining the meta-network, we consider building the sample-wise selector capable of jointly optimizing token number $N$ and model channels $C$. 
This selector, integral to the framework, should meet two paramount criteria: \emph{(1) lightweight} and \emph{(2) sample-effective}. 
The former necessitates minimal consumption of storage overheads and computational resources to circumvent parity with direct classification. 
Conversely, the latter indicates effective learning of decision-making processes to maximize the utilization of valid information in both the model parameters and the data, while minimizing resource consumption.
In this paper, we model the token and channel reduction process as a Markov decision process and employ PPO accordingly. 

\noindent \textbf{Formulation of Joint Token and Architecture Optimization.} 
To reduce the decision cost, we apply the selector within each Transformer group, consisting of every three Transformer blocks, to determine the token optimization (\ie, pruning, merging, or pruning-merging) ratio and the network structures for the subsequent blocks, tailored to each sample. 
Therefore, the selector is formulated as: 
\begin{equation}
    {\textbf{s}^{(l)}},~\textbf{t}^{(l)} = \mathcal{F}_{\texttt{Selector}} \left ({\mathbf{O}^{\left (l-1 \right)}} \right ),
    \label{_eq:selector}
\end{equation}
where $l$ is the group index, ${\textbf{t}^{(l)}}$ represents the token keep ratio for $l$-th Transformer group, ${\textbf{s}^{(l)}}$ denotes the structures decision of $l$-th Transformer group, and the $\mathbf{O}^{(l - 1)}$ is the feature extracted by the $(i-1)$-th Transformer group, representing the abstracted data information up to the current block in the ViT. According to previous works~\cite{liang2022evit,tome,dynamicvit}, \texttt{<CLS>} token, $\mathbf{Q}$ (query), $\mathbf{K}$ (key), $\mathbf{V}$ (value), and the output of Self-Attention can be used. 
For simplicity, we omit the notations of dimensions for these matrices. 

Token optimization consists of two steps: \emph{(1) token importance ranking} and \emph{(2) token optimization}. 
In the first step, tokens are sorted by their contributions to the task, so that a specific token optimization method can be applied in the second step according to the token keep ratio. 
For token importance ranking, as \texttt{<CLS>} progressively aggregates the task-specific (\eg, classification) global information, the inner product of class token \texttt{<CLS>} and other tokens reflect the importance of different tokens. 
Hence, we leverage this mechanism for token ranking to measure whether a token is important to the input samples. 
To get accurate informative information, we directly use the first MHSA layer $\texttt{MHSA}_{1}^{(l)}$ in the $l$-th Transformer group to extract the importance vector $\textbf{a}_{cls}^{(l)}$ for the output of last Transformer group $\mathbf{Y}^{(l-1)}$ to avoid an additional matrix multiplication: 
\begin{align}
  \mathbf{X}^{(l)} &= \texttt{Sort} \left( \texttt{MHSA}_{1}^{(l)} \left(\mathbf{Y}^{(l-1)}; \textbf{s}^{(l)} \right), \alpha_{cls}^{(l)} \right) \quad \text{where} \notag \\
  \alpha_{cls}^{(l)} &= \texttt{Softmax} \left( \frac{\mathbf{q}_{cls}^{(l)} \cdot \left( \mathbf{K}^{(l)} \right)^{\mathsf{T}}}{\sqrt{C^{(l)}}} \right) \mathbf{V}^{(l)},
  \label{_eq:attn_score}
\end{align}
where $\mathbf{q}_{cls}^{(l)}$ is the query of the class tokens. 
Therefore, ${\alpha}_{cls}^{(l)}$ of Eq. (\ref{_eq:attn_score}) is actually a vector of the output $\texttt{MHSA}_{1}^{(l)} (\mathbf{Y}^{(l-1)})$.  $\texttt{Sort}( \cdot )$ is the sorting function that can arrange tokens in descending order based on $\textbf{A}_{cls}^{l}$. 

After preprocessing the tokens, we consider three representative token reduction strategies to obtain the tokens for the remaining MHSA layers and MLP layers in $l$-th group: \emph{(i) pruning}, \emph{(ii) merging}, and \emph{(iii) pruning-then-merging}. 
For token pruning, unimportant tokens will be discarded for each sample according to ${\textbf{t}^{(l)}}$ \cite{liang2022evit}:
\begin{equation}
    {\mathbf{X}}^{(l)} = \mathbf{X}^{\left( l \right)} \left[: \texttt{round} \left( N^{\left( l \right)} \times {\textbf{t}^{\left(l \right)}} \right) , :\right ]. 
\end{equation}  

For token merging, the sorted tokens will be divided into two categories based on the token keep ratio ${\textbf{t}^{(l)}}$: important tokens $\mathbf{X}_\textbf{im}^{(l)}$ and unimportant tokens $\mathbf{X}_\textbf{un}^{(l)}$: 
\begin{align}
\small
    {\mathbf{X}}^{(l)}_\textbf{im} &= \mathbf{X}^{\left( l \right)} \left[: \texttt{round} \left( N^{\left( l \right)} \times {\textbf{t}^{\left(l \right)}} \right) , :\right ], \quad \text{and} \\ {\mathbf{X}}^{(l)}_\textbf{un} &= \mathbf{X}^{\left( l \right)} \left[ \texttt{round} \left( N^{\left( l \right)} \times {\textbf{t}^{\left(l \right)}} \right): , :\right ],
\end{align}  
Subsequently, each unimportant token $\mathbf{X}_\textbf{un}^i$ will be merged into an optimal important token $\mathbf{X}_\textbf{im}^{{j^*}}$ that is most similar to it, to formulate a new $\mathbf{X}^{(l)}$ for next layers:
\begin{equation}
\small
    \mathbf{X}^{(l)} = \{\mathbf{X}^{m}\}_{m=0}^M ,  \quad \text{where} \quad \mathbf{X}_\textbf{im}^{{m}} = \mathbf{X}_\textbf{im}^{m} + \mathbf{X}_\textbf{un}^n ,
\end{equation}
where $M=\texttt{round} \left( N^{\left( l \right)} \times {\textbf{t}^{\left(l \right)}} \right)$ represents the number of kept tokens, $m, n$ are the indexes which achieve maximal cosine similarity $\cos ({\bm {\theta}_{mn}})$, which is calculated by: 
\begin{equation}
    \cos ({\bm {\theta}_{mn}}) = \frac{{\mathbf{X}_\textbf{un}^m \cdot \mathbf{X}_\textbf{im}^n}}{{||\mathbf{X}_\textbf{un}^m|| \cdot ||\mathbf{X}_\textbf{im}^n||}}.
\end{equation}
For pruning and merging, we adopt the pruning-then-merging \cite{diffrate} scheme. Specifically, the token keep ratio is divided into a token pruning ratio $\textbf{t}_{prune}^{(l)}$ along with a token merging ratio $\textbf{t}_{merge}^{(l)}$, i.e., $\textbf{t}^{(l)} = \left \{\textbf{t}_{prune}^{(l)},\textbf{t}_{merge}^{(l)}\right \}$. 

After the token optimization, the remaining tokens will go through the latter Transformer blocks within this group, with the architectural decisions based on ${\textbf{s}^{(l)}}$. 

\noindent \textbf{Lightweight Selector Modeling.} 
For the proposed \sysname, it is crucial to construct a lightweight yet high-performing selector capable of sample-aware optimization for both model structure and tokens, ensuring accuracy and FLOPs within a large optimization space. 
However, it is non-trivial to learn the solution via conventional supervised learning, as it is difficult to collect substantial labeled data.
Considering the outstanding performance of RL in various decision-making tasks such as gaming \cite{lbc,reviewatari}, control \cite{lee_ddpg, lee_td3}, combinatorial optimization \cite{asp_solver}, and data augmentation \cite{gdi}, we decided to use the PPO algorithm to effectively optimize the selector in such a large decision space.
Among serious RL methods, PPO \cite{ppo} is one of the on-policy algorithms with relatively stable performance and wide applicability, which has even been applied to many popular language models to improve their effectiveness \cite{rlhf}. 

The training of PPO consists of two components: an actor network for generating the decisions for architectural optimization and token reduction, and a value network $\mathcal{V}$ for predicting the value of the current state. 
During the training process, the value network will evaluate the current state ${\mathbf{O}^{(l)}}$ first, then the actor network will generate the corresponding policy $\{{\textbf{s}^{(l)}},~\textbf{t}^{(l)}\}$ based on the evaluation to reach the states with higher value. 
During inference, only the actor network is used to serve as the selector $\mathcal{F}_{\tt Selector}$. 

The training process involves limiting the magnitude of each policy update to prevent drastic changes, by adopting a clipped loss function to control the difference between the new and old policies and optimizing this loss function to improve policy performance. 
Specifically, for the $l$-th Transformer group, we use GAE (Generalized Advantage Estimation) \cite{GAE} to calculate the advantage function, and the objective of the value network $\mathcal{V}$ is to predict the state value with utmost accuracy, thus we employ a loss function derived from Temporal-difference methods: 
\begin{equation}
\left\{ {\begin{array}{*{20}{l}}
  {\textbf{L}_V^{(l)} = {{\left( {{\textbf{r}^{(l)}} + \gamma \mathcal{V}({\mathbf{O}^{(l + 1)}}) - \mathcal{V}({\mathbf{O}^{(l)}})} \right)}^2} \big/ B} \\ 
  {\hat {\textbf{A}}_{GAE(\gamma ,\lambda )}^{(l)} = \sum\limits_{i = 0}^\infty  {{{\left( {\gamma \lambda } \right)}^i}\textbf{L}_\mathcal{V}^{(l + i)}} } 
\end{array}} \right. ,
\end{equation}
where $B$ is batch size, $\gamma$ is the discount factor, $\textbf{r}^{(l)}$ is the reward of $l$-th Transformer group, and $\mathbf{O}^{(l)}$ is the state information (the input of selector in Eq. (\ref{_eq:selector})) of the $l$-th Transformer group, $\hat {\textbf{A}}^{(l)}$ denotes the estimation of the advantage function. Then the optimal policies can be obtained by maximizing the rewards of the sequence: 
\begin{equation}
\begin{aligned}
    {{\textbf{L}}^{CLIP}}(\theta ) = &- {{\hat {\mathbb{E}}}^{(l)}}\left[ \min \left( {r^{(l)}}(\theta ){{\hat {\textbf{A}}}^{(l)}}, \Phi \right) \right. \\
    & \left. - \delta {{\mathbb{E}}_{{a^{(l)}} \sim \pi }}\left[ - \log \left( \pi ({{\textbf{a}}^{(l)}}|{\mathbf{O}^{(l)}}) \right) \right] \right], \\
    \text{where} \quad \Phi = & \texttt{Clip}\left( {r^{(l)}}(\theta ), 1 - \varepsilon, 1 + \varepsilon \right){{\hat {\textbf{A}}}^{(l)}},
\end{aligned}
\end{equation}
$\pi$ is the policy of the actor network $A$,  $\delta$ is the policy entropy ratio, which will facilitate PPO's exploration of the action space, ${r^{(l)}}(\theta )$ is the import sampling ratio of current policy and old policy:
\begin{equation}
    r^{(l)}(\theta) = \frac{\pi_\theta (\textbf{a}^{(l)} | \mathbf{O}^{(l)})}{\pi_{\theta_{\text{old}}} (\textbf{a}^{(l)} | \mathbf{O}^{(l)})}.
\end{equation}
Besides, to achieve better results, we optimized PPO with techniques such as Advantage Normalization \cite{Advantage_Normalization}, Gradient Clip, and Orthogonal Initialization. 

\begin{figure*}[t]
    \centering
    \includegraphics[width=1.0\textwidth]{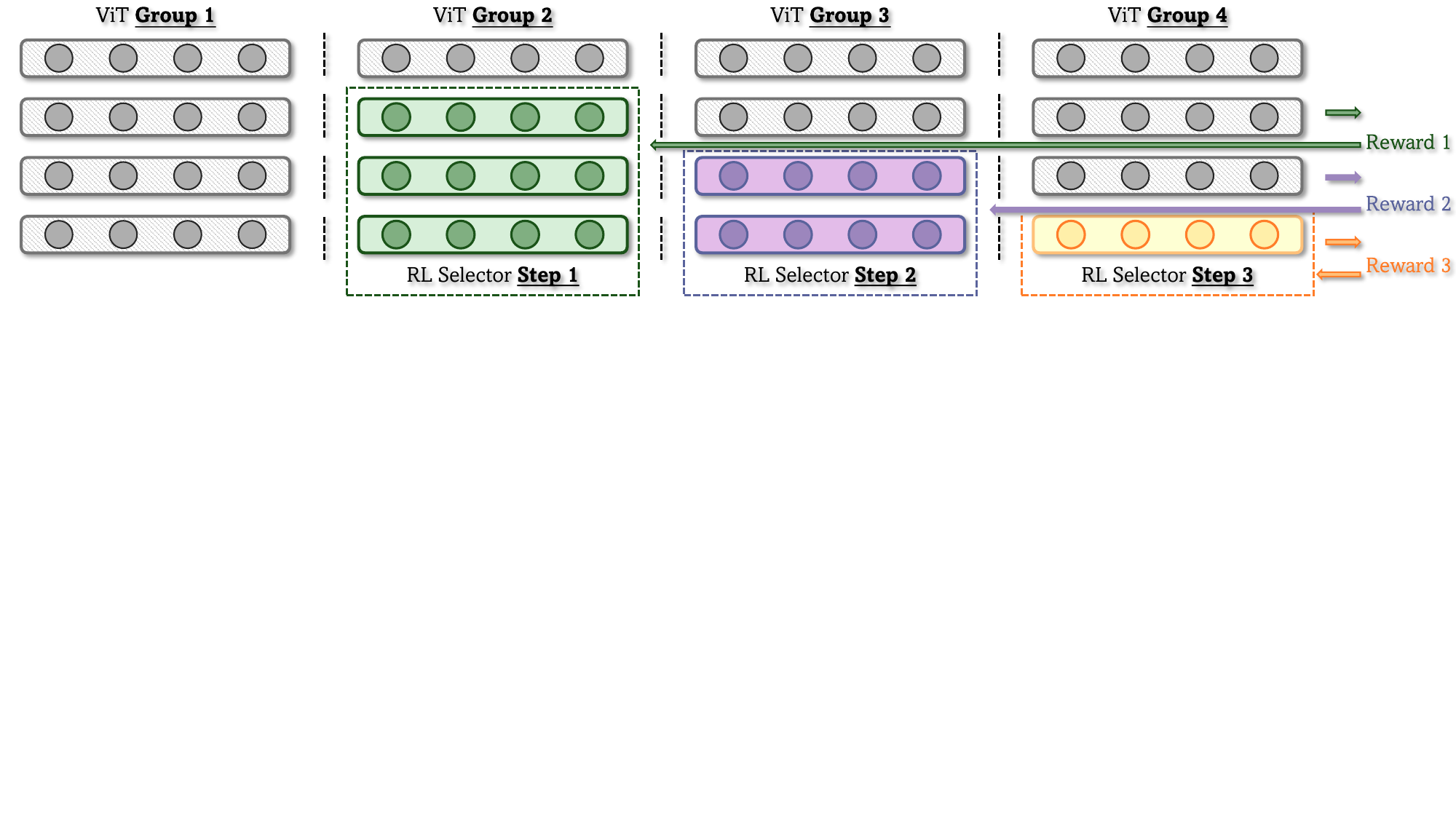}
    \caption{
    \textbf{The workflow of ``Result-to -Go".} This mechanism is only used for training the selector. To receive immediate feedback for each decision, the meta-network is divided into multiple groups. Initially, the meta-network is set to the maximum channel number for all groups. The selector then optimizes the model channels and tokens numbers for a single group at a time, allowing the meta-network to run to the end and obtain immediate feedback. Since the meta-network is fixed, its inference process can be viewed as a Markov decision process, allowing the selector to modify the structure of the meta-network groups one by one.
    }
    \label{result_to_go}
\end{figure*}

\noindent \textbf{Sequence model of ``Result-to-Go".} 
In general, it is crucial for RL algorithm to capture the impact of each decision on the final result during training \cite{nipsw23,rethink}. 
However, the ViT model encounters a \emph{delayed-return} scenario where the final result becomes available only after passing through all the ViT blocks during the sequence decision process, while it requires multiple selector decisions throughout this process. 
Inspired by \cite{decision_transformer, contextual_transformer}, we conduct a timely return model named ``Result-to-Go", which is shown in Fig \ref{result_to_go}. First of all, the maximum structural parameters and 100\% token keep ratio are assigned to the model. 
During the forward propagation process, the PPO selector will optimize the activated channels and useful tokens in the current ViT group, and the model will continue to the end to acquire the classification result without changing the parameters of other groups. 
Following this paradigm, the activated channels and the useful token numbers of each ViT group will be modified gradually and progressively. 
It is worth mentioning that the changes of model structures and token numbers are tailored to specific samples, rather than being applied at a coarse-grained task level.

In this way, although obtaining the result of a single layer-by-layer decision requires multiple rounds of forward propagation, we can get timely effects for each decision. 
During the training process, the parameters of ViT are held constant, and the paradigm of "Result-to-Go" is utilized to train the PPO. However, it will be disabled during inference to ensure model efficiency.

\noindent \textbf{Reward function.} During PPO training, the reward function varies for each sample to achieve a sample-wise selector. Specifically, it consists of three parts: \emph{(1) Top-1 accuracy reward ${\textbf{r}_{acc}}$}, \emph{(2) FLOPs penalty ${\textbf{r}_{f}}$} and \emph{(3) token optimization penalty ${\textbf{r}_{t}}$}. 
Although we have obtained the meta-network with high accuracy on ImageNet, the classification results of individual samples still vary between correct and incorrect. Providing feedback to the selector based on it may introduce some disturbance: even if PPO generates optimal structural parameters and token optimization rates, incorrect feedback may still occur due to limitations of model's classification level. In order to overcome it, we maintain two sets of results during training: the classification result $\textbf{y}_t$ of the supernet and the dynamically optimized classification result $\textbf{y}$ involving the selector. 
If $\textbf{y}$ is the same as $\textbf{y}_t$, then $\textbf{y}$ is set to 1. In this way, the selector can obtain smooth and accurate feedback. The total reward function is: 
\begin{equation}
    \textbf{r} = {\textbf{r}_{acc}} - {a_f}\textbf{f} - {a_t}{\textbf{t}_\textbf{r}}
    \quad \text{where} \quad 
    \textbf{r}_{\texttt{acc}} = \left\{
    \begin{array}{ll}
        \textbf{1}, & \texttt{if } \textbf{y} = \textbf{y}_t \\
        \textbf{r}_{\texttt{acc}}, & \texttt{otherwise}
    \end{array},
    \right. 
\label{loss function}
\end{equation}
where $a_f$ and $a_t$ represent the punishment factor of FLOPs ratio and token number ratio, respectively. They are utilized to control the trade-off between accuracy and FLOPs, and encourage token optimization. $\textbf{f}$ is the FLOPs of the model, ${\textbf{t}_\textbf{r}}$ represents the token retention ratio, ${\textbf{r}_f} =  - {a_f}\textbf{f}$ and ${\textbf{r}_t} =  - {a_t}{\textbf{t}_\textbf{r}}$. It is worth noting that ${\textbf{r}_t}$ is important. Due to the minor impact of modifying the network structure on accuracy, while token reduction has a greater impact on the result, PPO may tend to retain tokens, thereby falling into a local optimum. This phenomenon has also been observed in our experiments. 

\noindent \textbf{Actor function.} The actor function is constructed from the perspective of RL environment modeling. 
In terms of architectural optimization, the selector has a limited set of selections, namely the channels of MHSA and MLP layer, to choose from, making it a discrete decision problem. In terms of data optimization, the selector has to generate a token optimization ratio $\textbf{t} \in [0, 1]$, which leads to a continuous decision problem. 
Such a mixed-decision problem is hard to optimize because it significantly improves the complexity of PPO action space, we therefore transform the architectural optimization problem into a continuous decision problem. 

For $l$-th Transformer group, with each Transformer group has $K$ MHSA layers and $K$ MLP layers, the architectural decision involves $\mathbf{s}^{(l)} = \{{{s}^{(l)}_{\texttt{MLP}}}, {{s}^{(l)}_{\texttt{MHSA}}}\}$. 
Specifically, ${\mathbf{s}^{(l)}_{\texttt{MLP}}} = \{{{s}^{(l)}_{\texttt{MLP},k}}\}_{k=0}^K$ denotes the decided MLP ratios for the $K$ MLP layers in this Transformer group, where $\forall k \in K,  {{s}^{(l)}_{\texttt{MLP},k}} \in [0, 1]$, and 
${\mathbf{s}^{(l)}_{\texttt{MHSA}}} = \{{{s}^{(l)}_{\texttt{MHSA},k}} \}_{k=0}^K$
denotes the decided embedding dimension ratios for the $K$ MHSA layers in this Transformer group, $\forall k \in K,  {{s}^{(l)}_{\texttt{MHSA},k}} \in [0, 1]$. 
Suppose $E_\texttt{MHSA}$ and $E_\texttt{MLP}$ represent the number of architectural candidates of the MHSA and MLP layer, the index of the architectural parameters of this group can be calculated by: 
\begin{align} 
\small 
    {\mathbf{i}^{(l)}_{\texttt{MHSA}}} &= \left\{{ \texttt{round} \left( {s}^{(l)}_{\texttt{MHSA},k} * {E}_\texttt{MHSA} \right) } \right\}_{k=0}^K, \\
    {\mathbf{i}^{(l)}_{\texttt{MLP}}} &= \left\{{ \texttt{round} \left( {s}^{(l)}_{\texttt{MLP},k} * {E}_\texttt{MLP} \right) } \right\}_{k=0}^K. 
\end{align}
Accordingly, we can obtain the architectural parameters based on these indexes. 

The token $\textbf{t}^{(l)}$ represents the token pruning keep ratio, token merging keep ratio, or a combination of both, denoted as $\textbf{t}^{(l)} = \left \{\textbf{t}_{prune}^{(l)},\textbf{t}_{merge}^{(l)}\right \}$. Depending on the selected token optimization policies, it serves as a basis for conducting sample-specific token optimization.

\noindent \textbf{Masked Selector Training.} 
It is infeasible to train the selector on GPUs, due to diverse token numbers and embedding dimensions across different samples make the ViTs cannot perform batched parallel computations. 
Therefore, we leverage masking to enable parallel computation within existing frameworks for training the selector. 

Specifically, suppose $B$, $N$, and $C$ represent the batch size, maximum token number, and maximum embedding dimension, respectively, therefore $i \in [0,B]$, $j \in [0,N]$, and $k \in [0,C]$ denote the indices of batch size, number of tokens, and number of channels. The MLP Mask $\textbf{M}_{ijk}^L$ is applied to the MLP layers by padding zero to the channels beyond the selected embedding dimension for different samples to align the dimension, and Token Mask $\textbf{M}_{ijk}^{\mathsf{T}}$ add masks to tokens of different samples: 
\begin{equation}
\begin{split}
\textbf{M}_{ijk}^L = \begin{cases} 
1, & \texttt{if } k < {\textbf{d}_\textbf{e}}[i] \\ 
0, & \texttt{otherwise} 
\end{cases},
\end{split}
\quad
\begin{split}
\mathbf{X} = \mathbf{X} \odot \textbf{M}_{ijk}^L ,
\end{split}
\end{equation}

\begin{equation}
\begin{split}
\textbf{M}_{ijk}^{\mathsf{T}} = \begin{cases} 
1, & \texttt{if } j < {\textbf{d}_\textbf{t}}[i] \\ 
0, & \texttt{otherwise} 
\end{cases},
\end{split}
\quad
\begin{split}
\mathbf{X} = \mathbf{X} \odot \textbf{M}_{ijk}^{\mathsf{T}} ,
\end{split}
\end{equation}
where ${\textbf{d}_\textbf{e}}$ and $\textbf{d}_\textbf{t}$ represent vectors containing selected embedding dimensions and token numbers of samples, respectively, and $i$ denotes the sample number. $\odot$ represents the element-wise multiplication.

When it comes to Token Merging, the matrices $\mathbf{X}_\textbf{im}$ and $\mathbf{X}_\textbf{un}$, containing important and unimportant tokens respectively, can be constructed through Token Mask, where the masked tokens are set to $\infty$ instead of 0. Then the cosine similarity matrix is $\textbf{S}=\mathbf{X}_\textbf{un} \odot \mathbf{X}_\textbf{im}^{\mathsf{T}}$. After setting $\infty$ to $-\infty$, the useful information will be concentrated in the lower-left corner of matrix $\textbf{S}$, enabling sample-wise token merging based on it. This approach also relaxes the restriction of merging up to 50\% of the tokens by ToMe \cite{tome}. 

For LayerNorm which requires the computations of mean and standard deviation, we suppose the mask token matrix is $\mathbf{X}$ and the mask matrix is $\textbf{M}_{ijk}^L$. Initially, we perform mean filling for the parts of the token matrix that are masked, to mitigate the potential impact of mean calculation on the performance of the LayerNorm function. The valid sum value and the number of them can be calculated as follows: 
\begin{equation}
    {\mathbf{X}_{\text{mask}}}[i,j] = \sum\limits_{k = 1}^C {{\mathbf{X}_{ijk}}} ,\quad {\textbf{S}_X}[i,j] = \sum\limits_{k = 1}^C {\textbf{M}_{ijk}^L}. 
\end{equation}
The mean value will be filled into the token matrix:
\begin{align}
\small
    {\mathbf{X}_{\text{fill}}} = \mathbf{X} + (1 - {\textbf{M}^L}) \cdot {\mathbf{X}_{\text{mean}}} \quad \text{where}   \quad
     {\mathbf{X}_{\text{mean}}} = \frac{{{\mathbf{X}_{\text{mask}}}[i,j]}}{{{\textbf{S}_X}[i,j]}}. 
\end{align}

Next, the token matrix filled with mean value can be used to do sample-wise Layernorm:
\begin{equation}
    \mathbf{X} = \texttt{Layernorm} ({\mathbf{X}_{\text{fill}}}) \cdot \frac{{\sqrt N }}{{\sqrt {{\textbf{S}_X}} }}.
\end{equation}
\section{Experimental result}

In this section, we conduct extensive experiments to demonstrate the effectiveness of \sysname. 
All the experiments are conducted on ImageNet, and we report results for three different-sized models: ViT-Tiny, ViT-Small, and ViT-Base. 
Specifically, these models have roughly 1.2 GFLOPS, 5G FLOPs, and 20 GFLOPs, respectively. 
All experiments besides the selector training are performed on NVIDIA A100 GPUs with 40G memory with PyTorch training system, whereas we train the selector using a single NVIDIA RTX3090 GPU. 

\subsection{Experimental Settings.} 
\noindent \textbf{Meta-network.} 
The meta-networks are conducted based on the architecture of DeiT \cite{touvron2021training}, which consists of a total of 12 Transformer blocks. 
The embedding dimension and MLP dimension of the Attention layer are configurable. Besides, in order to further provide more selections for the PPO selector, we increase the structural complexity of the MLP by selecting multiples of the embedding dimension. 
The detailed model information can be found in Tab.~\ref{tab:supernet_setting}. 
Specifically, we train the models for 500 epochs using the AdamW optimizer and adopt mixed-precision training, and the first 20 epochs are used for warm-up with the learning rate set to $1 \times 10^{-6}$. 
We use the cosine learning rate scheduler for training. The initial learning rate is set to $5 \times 10^{-4}$, the decay rate is 0.1, and the minimum learning rate is $1 \times 10^{-5}$. 

\begin{table}[t]
    \caption{Setting of the Meta-network.}
    \centering
    \begin{threeparttable}          
    \scalebox{0.9}{
    \begin{tabular*}{\linewidth}{@{\extracolsep{\fill}}ccccc@{}}
        \toprule
        \textbf{Model} & \textbf{Embedding Dim}       & \textbf{MLP Ratio} & \textbf{Heads} & \textbf{Depth} \\ \midrule
        Tiny       & \{176, 192, 216, 240\}           & \{2, 4, 6\}            & 3                     & 12             \\
        Small       & \{320, 352, 368, 384, 400, 416\} & \{2, 3.5, 5\}          & 7                     & 12             \\
        Base      &  \{672, 696, 720, 744, 768\}          & \{2, 3.5, 5\}          & 12                    & 12             \\
        \bottomrule          
    \end{tabular*}
    }
    \begin{tablenotes}    
        \footnotesize              
        \item The embedding dimension and MLP dimension of the Attention block are set as optional structural dimensions, and the structural space is expanded by setting the MLP dimension option to be the multiple of the embedding dimension. 
      \end{tablenotes}            
    \end{threeparttable}       
    \label{tab:supernet_setting}
\end{table}

\noindent \textbf{PPO Selector.} 
To obtain accurate information for support decisions, we enable the selector after the first Transformer group. 
In other words, the token and architecture will not be changed during the inference of the first Transformer group. 
For each Transformer group $l \in \{2, ... ,L\}$ that is to be decided, the outputs of the selector include the token keep ratio ${\textbf{t}^{(l)}}$, and the architectural decision $\textbf{s}^{(l)} = \{{\textbf{s}^{(l)}_{\texttt{MLP}}}, {\textbf{s}^{(l)}_{\texttt{MSA}}}\}$, where ${\textbf{s}^{(l)}_{\texttt{MLP}}}$ represents the MLP expansion ratios and ${\textbf{s}^{(l)}_{\texttt{MSA}}}$ represents the embedding dimension of MSA layer. 
The ${\textbf{t}^{(l)}} \in [0,1]$ will be directly used to optimize the tokens of the current group. 

The actor network and critic network of the PPO selector each consist of 3 fully connected layers, with a state dimension of 197 and a hidden dimension of 256. 
The actor dimension depends on token optimization strategies: it is set to 7 for pruning and merging, 8 for pruning-then-merging. 
The learning rate of actor network and critic network are $1 \times 10^{-4}$ and $5 \times 10^{-3}$, respectively. 
The token punish ratio $a_t$ is 0.2. 
The FLOPs punish ratio $a_f$ is determined by the peak FLOPs of the meta-network, with a punishment range of [0.2, 0.6]. For different model scales, the specific ranges are set as follows: [0.02, 0.04] for tiny models, [0.008, 0.015] for small models, and [0.002, 0.006] for base models. 
The interval is 0.005 for both the tiny and base models, and 0.002 for the small model.
Besides, the selector is trained for 1 epoch using $50000$ images sampled from the training dataset. 
During inference, decisions can be averaged directly over the batch dimension without performance degradation. 

\noindent \textbf{Finetuning.} 
We follow the fine-tuning settings of DynamicViT~\cite{dynamicvit}. 
Specifically, we finetune the meta-network with 30 epochs using the cosine learning rate scheduler and do not perform warm-up, we use an external CNN teacher to further improve the performance. 
During fine-tuning, we freeze the PPO selector. 
The initial learning rate is set to $2 \times 10^{-5}$, the minimize learning rate is $2 \times 10^{-6}$, the weight decay is $1 \times 10^{-6}$. The mixup is disabled to improve the convergence. 

\begin{table*}[htbp]
  \centering
  \caption{The main results of \sysname.}
  \begin{threeparttable}
    \begin{tabular*}{\linewidth}{@{\extracolsep{\fill}}ccccccc@{}}
    \toprule
     \textbf{Model}     & \textbf{Method} & \textbf{Top-1 Acc. (\%)} & \textbf{FLOPs (G)} & \textbf{Token Keep Rate} &\textbf{\makecell[c]{Architectural \\ Optimization}} & \textbf{\makecell[c]{Token Optimization \\ Strategy}} \\
    \midrule
    \multirow{13}[4]{*}{ViT-Tiny} & DeiT-T & 72.20\% & 1.20  & 100\% & - & N/A  \\
\cmidrule{2-7}   
          & SAViT \cite{savit} & 70.72\% \down{1.48} & 0.90 \down{25\%} &   N/A  & \checkmark & N/A \\
          & UPDP \cite{updp}  & 70.12\% \down{2.08} & 0.90 \down{25\%} &  N/A    & \checkmark & N/A \\
          & A-ViT \cite{A-vit} & 71.00\% \down{1.20} & 0.80 \down{33\%}  & - & $\times$ & Token Pruning \\
          & DynamicViT~\cite{dynamicvit} & 70.90\% \down{1.30} & 0.90 \down{25\%}  & \tokenfixed{70.00\%} & $\times$ & Token Pruning \\
          & S$^2$ViTE \cite{S2ViTE} & 70.12\% \down{2.08} & 0.90 \down{25\%}  & \tokenfixed{30.00\%}      & $\times$  & Token Pruning \\
          & SPViT~\cite{kong2022spvit} & 72.10\% \down{0.10} & 0.90 \down{25\%}  & \tokenfixed{34.30\%}  
          & $\times$ & Token Pruning \\
          & Evo-ViT \cite{Evo_vit} & 72.00\% \down{0.20} & 0.73 \down{39\%}  & \tokenfixed{25.00\%} & $\times$ & Token Pruning \\
          & ToMe \cite{tome}  & 71.27\% \down{0.93} & 0.90 \down{25\%} & \tokenfixed{70.00\%}  & $\times$ &Token Merging \\
          \rowcolor{yellow!25} & Ours  & \textbf{72.38\%} \up{\textbf{0.18}} & 0.87 \down{28\%}  & \tokenlearned{25.00\%}  & \checkmark & Token Pruning \\
          \rowcolor{yellow!25} & Ours  & \textbf{72.81\%} \up{\textbf{0.61}} & 0.96 \down{20\%}  & \tokenlearned{53.00\%}  & \checkmark & Token Merging \\
          \rowcolor{yellow!25} & Ours  & \textbf{73.31\%} \up{\textbf{1.11}} & 0.87 \down{28\%}  & \tokenlearned{33.00\%}  & \checkmark & Token P + M \\
    \midrule
    \multirow{21}[4]{*}{ViT-Small} & DeiT-S & 79.90\% & 4.70  & 100\% & - &N/A  \\
    \cmidrule{2-7}
          & A-ViT \cite{A-vit} & 78.60\% \down{1.30} & 3.60 \down{23\%}  & - & $\times$ & Token Pruning \\
          & DynamicViT~\cite{dynamicvit}& 79.32\% \down{0.58} & 2.90 \down{38\%}  & \tokenfixed{34.30\%}  & $\times$ & Token Pruning \\
          & Evo-ViT \cite{Evo_vit} & 79.40\% \down{0.50} & 3.00 \down{36\%}  & \tokenfixed{25.00\%} & $\times$ &Token Pruning \\
          & SPViT~\cite{kong2022spvit} & 79.80\% \down{0.10} & 3.86 \down{18\%}  & \tokenfixed{34.30\%}  &$\times$ &Token Pruning \\
          & EViT~\cite{liang2022evit}  & 79.50\% \down{0.40} & 3.00 \down{36\%}  & \tokenfixed{34.30\%}  &$\times$ &Token Pruning \\
          & S$^2$ViTE \cite{S2ViTE} & 79.22\% \down{0.68} & 3.20 \down{32\%}  & \tokenfixed{40.00\%}      & $\times$ &Token Pruning  \\
          & GTP-ViT~\cite{xu2024gtp} & 78.80\% \down{1.10} & 2.90 \down{38\%} & \tokenfixed{14.29\%} & $\times$ &Token Merging \\
          & ToMe \cite{tome} & 78.85\% \down{1.05} & 2.67 \down{43\%}   & \tokenfixed{27.00\%}  & $\times$ &Token Merging \\
          & DiffRate~\cite{diffrate} & 79.47\% \down{0.43} & 2.85 \down{39\%} & \tokenlearned{48.73\%} & $\times$ &Token P + M \\
          & LTMP~\cite{ltmp} & 79.30\% \down{0.60} & 2.70 \down{43\%} & - & $\times$ &Token P + M \\
          \rowcolor{yellow!25} & Ours  & \textbf{80.02\%} \up{\textbf{0.12}} & 2.75 \down{41\%}  & \tokenlearned{36.00\%}  & \checkmark &Token Pruning \\
          \rowcolor{yellow!25} & Ours  & \textbf{80.17\%} \up{\textbf{0.27}} & 2.85 \down{39\%}  & \tokenlearned{38.00\%}  & \checkmark &Token Merging \\
          \rowcolor{yellow!25} & Ours  & \textbf{80.19\%} \up{\textbf{0.29}} & 2.96 \down{37\%}  & \tokenlearned{33.00\%}  & \checkmark &Token P + M \\
             
    \cdashline{2-7}
          & ViT-Slimming \cite{chavan2022vision} & 77.90\% \down{2.00} & 2.30 \down{51\%}  & N/A   & \checkmark & N/A \\
          & X-Pruner \cite{chavan2022vision} & 78.93\% \down{0.97} & 2.40 \down{49\%}  & N/A  & \checkmark & N/A \\
          & EViT~\cite{liang2022evit} & 78.50\% \down{1.40} & 2.30 \down{51\%}  & \tokenfixed{12.50\%} & $\times$ &Token Pruning \\
          & DynamicViT \cite{dynamicvit} & 77.50\% \down{2.40} & 2.20 \down{53\%}  & \tokenfixed{50.00\%}  & $\times$  &Token Pruning \\
          & GTP-ViT~\cite{xu2024gtp} & 78.30\% \down{1.60} & 2.60 \down{45\%} & \tokenfixed{14.29\%} & $\times$ &Token Merging \\
          & Diffrate \cite{diffrate} & 78.81\% \down{1.09} & 2.40 \down{49\%}  & \tokenlearned{33.50\%}  & $\times$ &Token P + M \\
          \rowcolor{yellow!25} & Ours  & \textbf{79.40\%} \down{0.50} & 2.28 \down{53\%}   & \tokenlearned{13.00\%}  & \checkmark &Token Pruning \\
           \rowcolor{yellow!25} & Ours  & \textbf{79.98\%} \up{\textbf{0.08}} & 2.38 \down{49\%}  & \tokenlearned{18.00\%}  & \checkmark &Token Merging \\
           \rowcolor{yellow!25} & Ours  & \textbf{79.98\%} \up{\textbf{0.08}} & 2.55 \down{46\%}  & \tokenlearned{33.00\%}  & \checkmark &Token P + M \\
    \midrule
    \multirow{12}[4]{*}{ViT-Base} & DeiT-B & 81.80\% & 18.00  & 100\% & - & N/A \\
\cmidrule{2-7}
          & AS-DeiT-B \cite{liu2022adaptive}       & 81.40\% \down{0.40}   &  11.20 \down{38\%}  &     \tokenfixed{34.30\%}  & $\times$ &Token Pruning \\
          & DynamicViT~\cite{dynamicvit} & 81.43\% \down{0.37} & 11.40 \down{37\%} & \tokenfixed{34.30\%} & $\times$ &Token Pruning \\
          & EViT~\cite{liang2022evit} & 80.00\% \down{1.80} & 8.70 \down{52\%} & \tokenfixed{12.50\%}  & $\times$ &Token Pruning \\
          & EViT~\cite{liang2022evit} & 81.30\% \down{0.50} & 11.50 \down{36\%} & \tokenfixed{34.30\%} & $\times$ &Token Pruning \\
          & Evo-ViT \cite{Evo_vit} & 81.30\% \down{0.50} & 11.31 \down{37\%}  & \tokenfixed{25.00\%} & $\times$ &Token Pruning \\
          & GTP-ViT~\cite{xu2024gtp} & 80.70\% \down{1.10} & 11.00 \down{38\%} & \tokenfixed{32.65\%} & $\times$ &Token Merging \\
          & ToMe~\cite{tome} & 80.29\% \down{1.51} & 10.57 \down{41\%} & \tokenfixed{27.00\%} & $\times$ & Token Merging \\
          & Diffrate~\cite{diffrate} & 79.73\% \down{2.07} & 9.42 \down{48\%} & \tokenlearned{30.46\%} & $\times$ &Token P + M \\
          & Diffrate~\cite{diffrate} & 81.27\% \down{0.53} & 11.61 \down{36\%} & \tokenlearned{47.72\%}  & $\times$ &Token P + M \\
          & LTMP~\cite{ltmp} & 78.80\% \down{3.00} & 9.00 \down{50\%} & - & $\times$ &Token P + M \\
          \rowcolor{yellow!25} & Ours  & \textbf{81.49\%} \down{0.31} & 10.90 \down{39\%}  & \tokenlearned{51.00\%}  & \checkmark &Token Pruning \\
          \rowcolor{yellow!25} & Ours  & \textbf{81.77\%} \down{0.03} & 10.59 \down{41\%}  & \tokenlearned{65.00\%}  & \checkmark &Token Merging \\
          \rowcolor{yellow!25} & Ours  & \textbf{81.52\%} \down{0.28} & 9.54 \down{47\%}  & \tokenlearned{67.00\%}  & \checkmark &Token P + M \\
    \bottomrule
    \end{tabular*}%
    \begin{tablenotes}    
        \footnotesize              
        \item We present the performance comparison of \sysname~with various SOTA methods across three model sizes: Tiny, Small, and Base, with the results of three different token optimization methods: Pruning, Merging, and pruning-then-merging (Token P+M). 
      \end{tablenotes}            
    \end{threeparttable} 
  \label{main_result}%
\end{table*}%

\subsection{Main results}
Tab.~\ref{main_result} shows the results of the \sysname~across three different-sized models.
The Top-1 accuracy, FLOPs, and token keep rate are reported, along with comparative analyses against other methods. 
Note that the token keep rate refers to the percentage of tokens retained after three rounds of optimization. 
One can see that \sysname~significantly reduces the FLOPs while maintaining exceptionally high accuracy. 
Under the same FLOPs constraints, \sysname~outperforms existing lightweight SOTA models. 
Notably, for ViT-Tiny and ViT-Small, our models even slightly surpass the accuracy of original models, showcasing that reducing the redundancies of ViTs can improve their generalization. 
Specifically, for ViT-Tiny, our model achieves approximately 1\% higher Top-1 accuracy than the base model while reducing FLOPs by about 30\%. 
For ViT-Small, with FLOPs reductions of approximately 40\% and 50\%, the Top-1 accuracy surpasses the base model by about 0.3\% and 0.1\%, respectively. 
For ViT-Base, with a reduction of approximately 40\% in FLOPs, the Top-1 accuracy is only 0.03\% lower than that of the base model. 

Overall, \sysname~is an efficient sample-wise inference method that optimizes both model structural dimensions and data dimensions simultaneously, enabling optimal results with minimal FLOPs. 
Moreover, by adaptively optimizing each sample through data optimization methods such as token pruning, token merging, token pruning-then-merging, and structural optimization methods such as channel selection, we can significantly reduce the model size while maintaining or even improving the model's accuracy. 
On the one hand, the optimization of both the model structure and data dimensions by \sysname~contributes significantly. On the other hand, the powerful adaptive decision-making capability of the PPO selector ensures that even the simplest optimization methods can be effectively applied to each sample, thus avoiding the need to use more complex lightweight techniques such as token clustering. 

\begin{table}[t]
  \centering
  \caption{The Details of PRANCE}
  \begin{threeparttable}
  \scalebox{0.9}{
    \begin{tabular}{ccccc}
    \toprule
    \textbf{Model} & \textbf{Top-1 Acc. (\%)} & \textbf{FLOPs(G)} & \textbf{\makecell[c]{Token Keep \\ Rate}} & \textbf{Type} \\
    \midrule
    DeiT-Tiny & 72.20\% & 1.20  & 100\% &  \\
    \midrule
    \multirow{10}[6]{*}{\makecell[c]{PRANCE\\(Tiny)}} 
          & 72.38\% \up{\textbf{0.18}} & 0.87 \down{28\%}  & 25\%  & Pruning \\
          & 73.07\% \up{\textbf{0.87}} & 0.97 \down{19\%}  & 25\%  & Pruning \\
          & 73.55\% \up{\textbf{1.35}} & 1.07 \down{11\%} & 60\%  & Pruning \\
\cmidrule{2-5}          
          & 71.39\% \down{0.81} & 0.87 \down{28\%}  & 51\%  & Merging \\
          & 72.81\% \up{\textbf{0.61}} & 0.96 \down{20\%}  & 53\%  & Merging \\
          & 73.05\% \up{\textbf{0.85}} & 1.03 \down{14\%} & 53\%  & Merging \\
\cmidrule{2-5}          
          & 72.36\% \up{\textbf{0.16}} & 0.71 \down{41\%}  & 8\%   & P+M \\
          & 73.31\% \up{\textbf{1.11}} & 0.87 \down{28\%} & 33\%  & P+M \\
          & 73.77\% \up{\textbf{1.57}} & 0.94 \down{22\%} & 41\%  & P+M \\
          & 74.41\% \up{\textbf{2.21}} & 1.03 \down{14\%} & 33\%  & P+M \\
    \midrule
    DeiT-Small & 79.90\% & 4.70  & 100\% &  \\
    \midrule
    \multirow{16}[6]{*}{\makecell[c]{PRANCE\\(Small)}} 
          & 78.93\% \down{0.97} & 2.07 \down{56\%}  & 11\%  & Pruning \\
          & 79.40\% \down{0.50} & 2.28 \down{51\%}  & 13\%  & Pruning \\
          & 79.59\% \down{0.31} & 2.59 \down{45\%} & 30\%  & Pruning \\
          & 80.12\% \up{\textbf{0.22}} & 2.84 \down{40\%} & 36\%  & Pruning \\
          & 80.25\% \up{\textbf{0.35}} & 3.17 \down{33\%} & 36\%  & Pruning \\
\cmidrule{2-5}          
          & 79.29\% \down{0.61} & 1.96 \down{58\%} & 18\%  & Merging \\
          & 79.98\% \up{\textbf{0.08}} & 2.38 \down{49\%} & 18\%  & Merging \\
          & 80.01\% \up{\textbf{0.11}} & 2.40 \down{49\%} & 18\%  & Merging \\
          & 80.05\% \up{\textbf{0.15}} & 2.58 \down{45\%} & 38\%  & Merging \\
          & 80.17\% \up{\textbf{0.27}} & 2.85 \down{39\%} & 38\%  & Merging \\
          & 80.21\% \up{\textbf{0.31}} & 3.05 \down{35\%} & 38\%  & Merging \\
\cmidrule{2-5}          
          & 79.42\% \down{0.48} & 2.10 \down{55\%} & 29\%  & P+M \\
          & 79.71\% \down{0.19} & 2.30 \down{51\%} & 29\%  & P+M \\
          & 79.98\% \up{\textbf{0.08}} & 2.55 \down{46\%} & 33\%  & P + M \\
          & 80.06\% \up{\textbf{0.16}} & 2.61 \down{44\%} & 33\%  & P + M \\
          & 80.15\% \up{\textbf{0.25}} & 3.25 \down{31\%} & 33\%  & P + M \\
    \midrule
    DeiT-Base & 81.80\% & 18.00  & 100\% &  \\
    \midrule
    \multirow{10}[6]{*}{\makecell[c]{PRANCE\\(Base)}} 
          & 81.29\% \down{0.51} & 9.60  \down{47\%} & 51\%  & Pruning \\
          & 81.43\% \down{0.37} & 10.41 \down{42\%} & 51\%  & Pruning \\
          & 81.49\% \down{0.31} & 11.34 \down{37\%} & 56\%  & Pruning \\
\cmidrule{2-5}          
          & 81.29\% \down{0.51} & 7.09 \down{61\%} & 36\%  & Merging \\
          & 81.42\% \down{0.38} & 7.64 \down{58\%} & 36\%  & Merging \\
          & 81.51\% \down{0.29} & 9.30 \down{48\%} & 7\%   & Merging \\
          & 81.77\% \down{0.03} & 10.59 \down{41\%} & 65\%  & Merging \\
\cmidrule{2-5}          
          & 80.24\% \down{1.56} & 7.43 \down{59\%}   & 27\%  & P + M \\
          & 81.23\% \down{0.57} & 8.20 \down{54\%}   & 47\%  & P + M \\
          & 81.52\% \down{0.28} & 9.54 \down{47\%}   & 67\%  & P + M \\
          & 81.62\% \down{0.18} & 11.66 \down{35\%}  & 73\%  & P + M \\
    \bottomrule
    \end{tabular}%
    }
    \begin{tablenotes}    
        \footnotesize              
        \item 
        \begin{minipage}[t]{0.46\textwidth}
        The results of \sysname~in jointly optimizing the model structure with three token optimization strategies: Pruning, Merging, and pruning-then-merging, across three different-sized models, including ViT-Tiny, ViT-Small, and ViT-Base. Besides, for each optimization scheme, results under multiple FLOPs constraints are reported to demonstrate the effectiveness of \sysname. 
        
        \end{minipage}
      \end{tablenotes}
    \end{threeparttable}
  \label{prance_detail_results}%
\end{table}%

Tab.~\ref{prance_detail_results} presents the results of \sysname~employing different token optimization strategies across various model settings. 
The results of each token optimization strategy at different optimization levels are reported. 
First of all, for models of various scales, \sysname~employing pruning, merging, and pruning-then-merging optimization strategies can maintain high accuracy with significantly reduced FLOPs. 
These results significantly surpass various lightweight SOTA models and even meet or exceed the performance of DeiT, demonstrating the powerful lightweight capability of our model and its excellent performance across different compression levels. Moreover, the results show that the three token optimization methods—pruning, merging, and pruning-then-merging—can all achieve excellent performance. This is different from the conclusions of more complex token optimization methods. With the help of the PPO selector, our model achieves a more reasonable token optimization ratio for different samples, even though it only uses sample token optimization methods. 

Meanwhile, token merging outperforms pruning at extremely low FLOPs. For instance, for ViT-Small with 1.96G FLOPs, the accuracy can still reach 79.29\%, for ViT-Base with 7.09G FLOPs, the accuracy can still reach 81.29\%. 
From the perspective of joint optimization of data and model structure, the token keep rate is the result of data dimension optimization, while the model channels represent the structural dimension optimization. For models of the same scale, the effect of data optimization and model optimization are coupled: increasing data or enhancing channel dimensions both improve accuracy, and there is a complementary relationship between them.
For example, \sysname~at the Base scale can achieve an accuracy of 81.51\% by retaining only 8\% of the tokens, or it can also achieve a similar accuracy of 81.49\% by retaining 56\% of the tokens.
What's more, although \sysname~can surpass DeiT in accuracy, the margin by which \sysname~exceeds DeiT decreases as the scale of DeiT increases. And \sysname~struggles to surpass DeiT in accuracy at the base scale.
This indicates that for ImageNet, the model's scale transitions from being insufficient to becoming overly large. At the Base scale, the training data is insufficient for the model to learn more.

\begin{table}[t]
    \centering
    \caption{The result of one-step decision and multi-step decision.}
    \begin{threeparttable}          
    \scalebox{0.9}{
        \begin{tabular}{ccccc}
            \toprule
            \textbf{Decision Type} & \textbf{Model} & \textbf{Top-1 Acc. (\%)} & \textbf{FLOPs(G)} & \textbf{\makecell[c]{Token Keep \\ Rate}} \\
            \midrule
            \multirow{3}[2]{*}{\textbf{Once}} & Tiny  & 62.44\% & 1.02  & 30\% \textcolor{blue}{([0.5, 1])} \\
                  & Small & 64.00\% & 3.18  & 36\% \textcolor{blue}{([0.5, 1])} \\
                  & Base  & 75.05\% & 10.9  & 42\% \textcolor{blue}{([0.5, 1])} \\
            \midrule
            \multirow{3}[2]{*}{\textbf{Step by step}} & Tiny  & 73.07\% & 0.97  & 25\% \textcolor{blue}{([0, 1])} \\
                  & Small & 80.12\% & 2.84  & 36\% \textcolor{blue}{([0, 1])} \\
                  & Base  & 81.43\% & 10.41 & 51\% \textcolor{blue}{([0, 1])} \\
            \bottomrule
        \end{tabular}%
        }
        \begin{tablenotes} 
        \begin{minipage}[t]{0.46\textwidth}
            \footnotesize              
            \item The intervals marked in blue indicate the range of token keep ratios for single-step decisions. Compared with one-step solution, the multi-step apporach can better capture the trade-off between accuracy with model structure and tokens. 
        \end{minipage}
        \end{tablenotes}            
    \end{threeparttable}     
    \label{once_vs_multi}%
\end{table}

\subsection{Analysis}
\noindent \textbf{One-step selection VS multi-step selection.} 
For the sample-wise joint optimization problem of data and model structure, the most intuitive approach is to generate all necessary parameters for lightweight inference at once, based on the initial samples. 
However, this approach treats ViT as a black box and constructs a substantially large decision space with up to around $10^{14}$ combinations. 
Another way is to follow the paradigm of RL by modeling the problem as a Markov process.
\sysname~divides ViT into multiple groups and allowing the PPO selector to make decisions in stages.
In this way, the action space is reduced to just 7 or 8 dimensions, which not only significantly reduces the difficulty of joint optimization but also helps the selector grasp the token features at different stages of the inference process.

Tab.~\ref{once_vs_multi} shows the results of the above two model ways.
The one-step selector fails to effectively learn the importance of tokens and model structure, resulting in poor performance. Conversely, the multi-step selector, modeled based on \sysname, can better capture the trade-off between accuracy and model structure with tokens, leading to better accuracy. 
Specifically, to avoid performance collapse, the token keep ratio for single-step decisions is set to [0.5, 1], while the \sysname~ranges from [0, 1].

\begin{figure*}[t]
    \centering
    \includegraphics[scale=0.19]{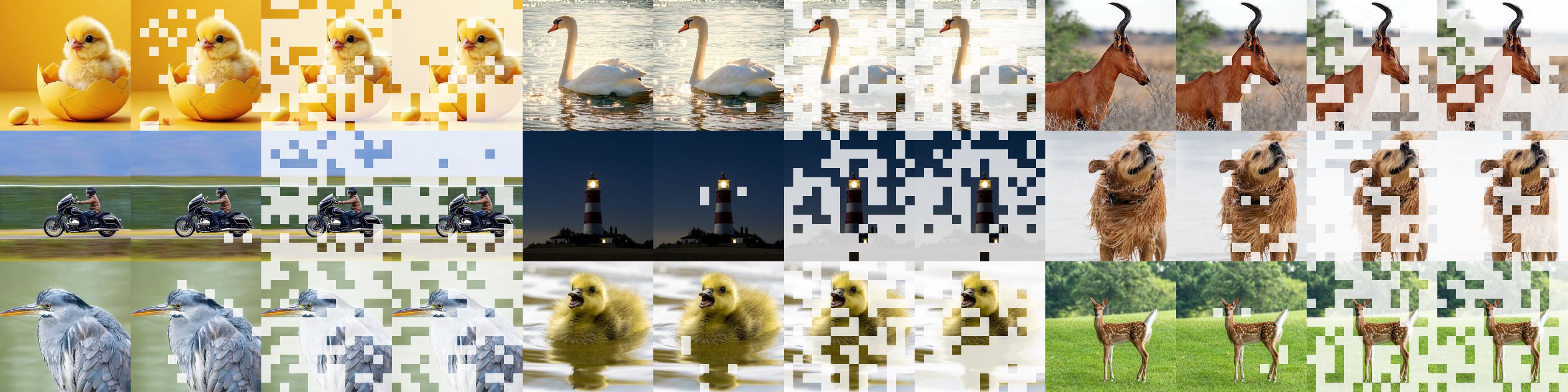}
    \caption{
    \textbf{Visualization of token pruning in different transformer groups.}
    \sysname~effectively identifies and retains important tokens while removing unimportant ones to reduce the complexity of ViTs. Besides, our framrwork prefers to retain tokens in the early stages and optimize a large number of low-information tokens in the later stages. 
    }
    \label{pruning_vis}
\end{figure*}

\begin{figure*}[t]
    \centering
    \includegraphics[scale=0.19]{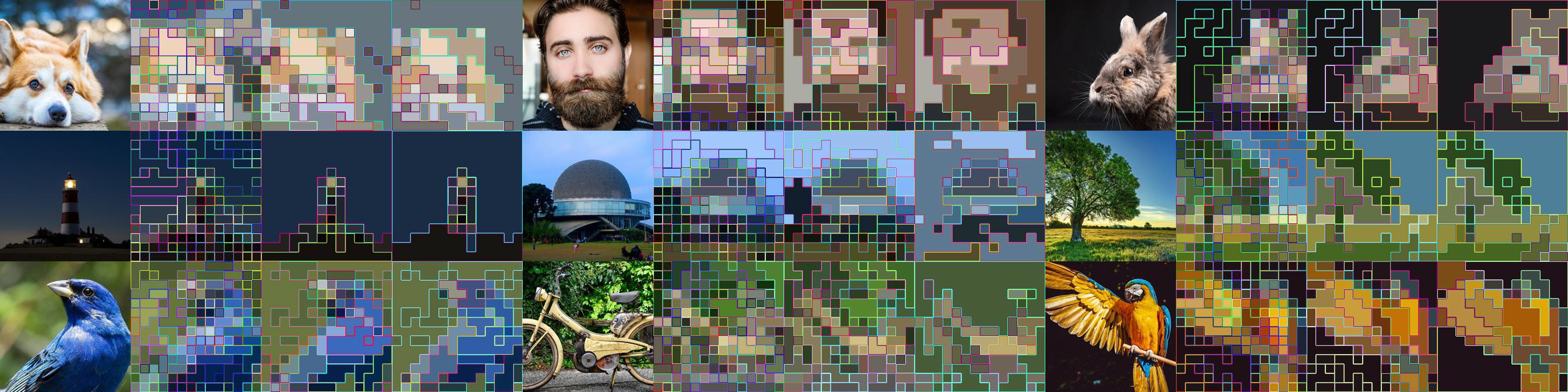}
    \caption{
    \textbf{Visualization of token merging in different transformer groups.} \sysname~can merge similar tokens based on their importance, retaining tokens with higher information. 
    }
    \label{merging_vis}
\end{figure*}

\begin{figure*}[t]
    \centering
    \includegraphics[scale=0.19]{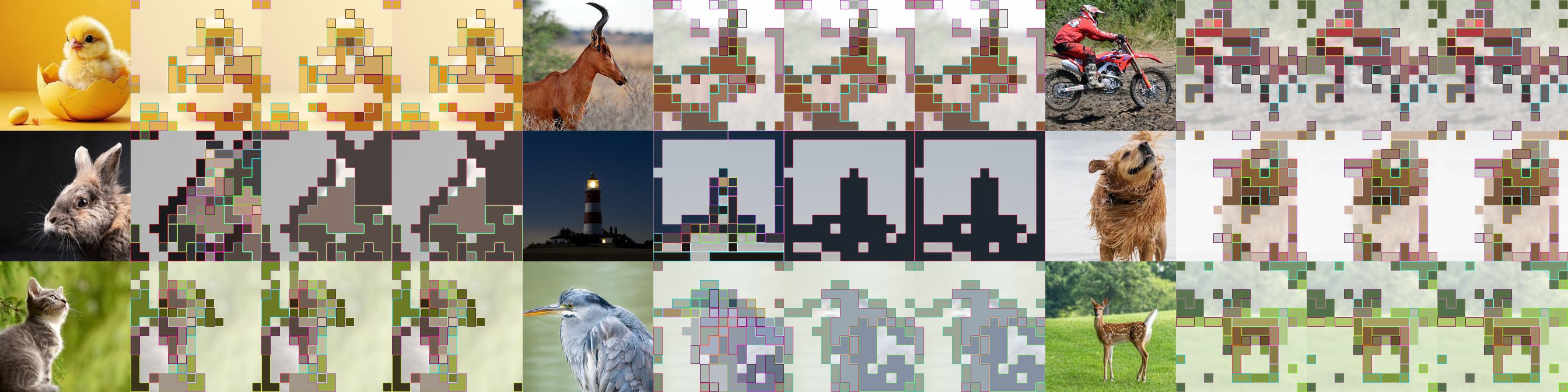}
    \caption{
    \textbf{Visualization of token pruning-then-merging (P+M) in different transformer groups.}
    The light, translucent parts represent the pruned tokens, while the colored blocks represent the merged tokens.
    \sysname~can prune the least informative background tokens based on the complexity of the image, then merge tokens with less information, and retain tokens with higher information content.
    }
    \label{pruning_merging_vis}
\end{figure*}

\begin{figure*}[!t]
    \centering
    \subfloat[Dimensional Distribution of Embedding Channels of the Model for the "Bird" Category.]{\includegraphics[width=1.5in]{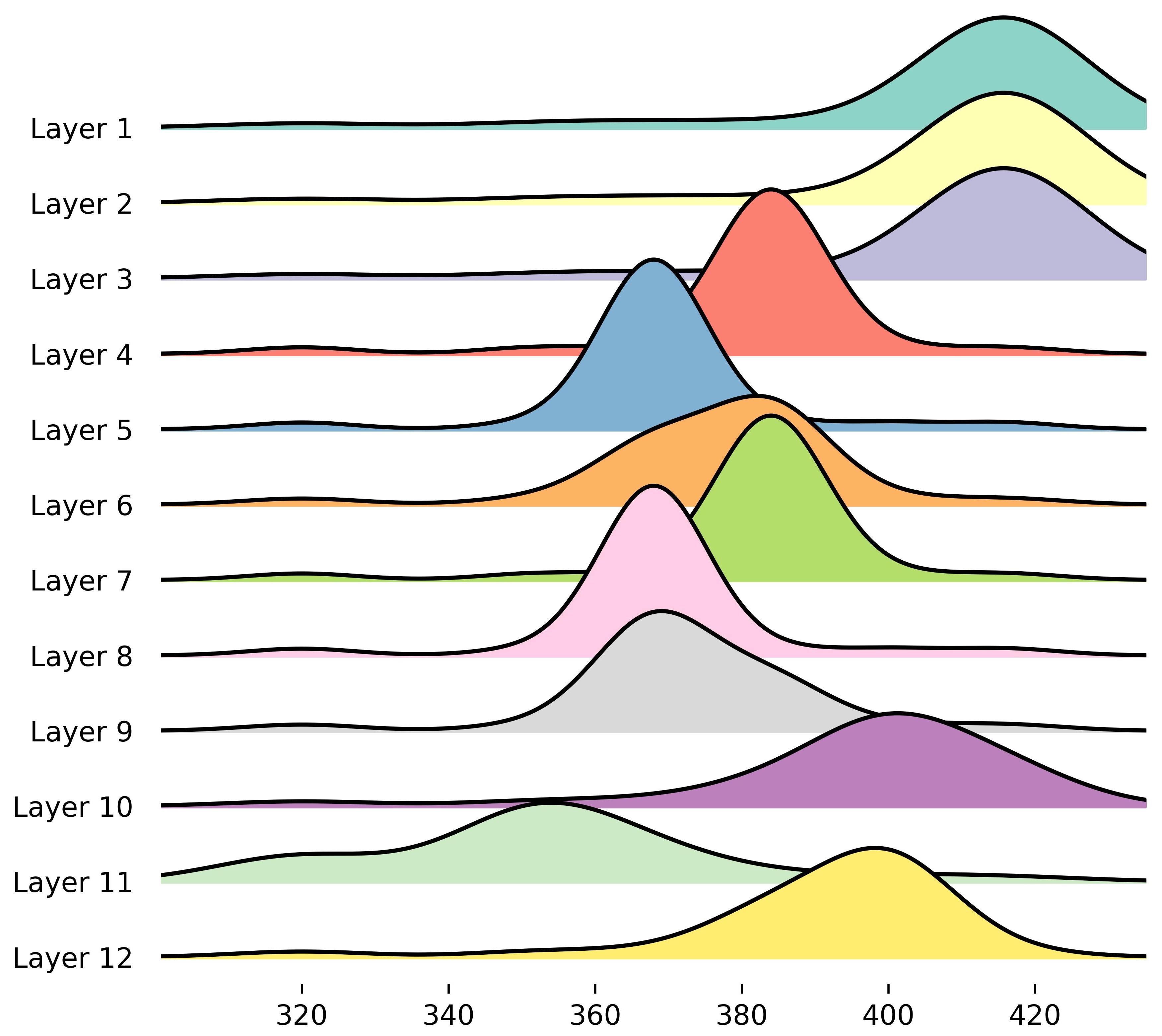}%
    \label{figure1}}
    \hfil
    \subfloat[Dimensional Distribution of MLP of the Model for the "Bird" Category.]
    {\includegraphics[width=1.5in]{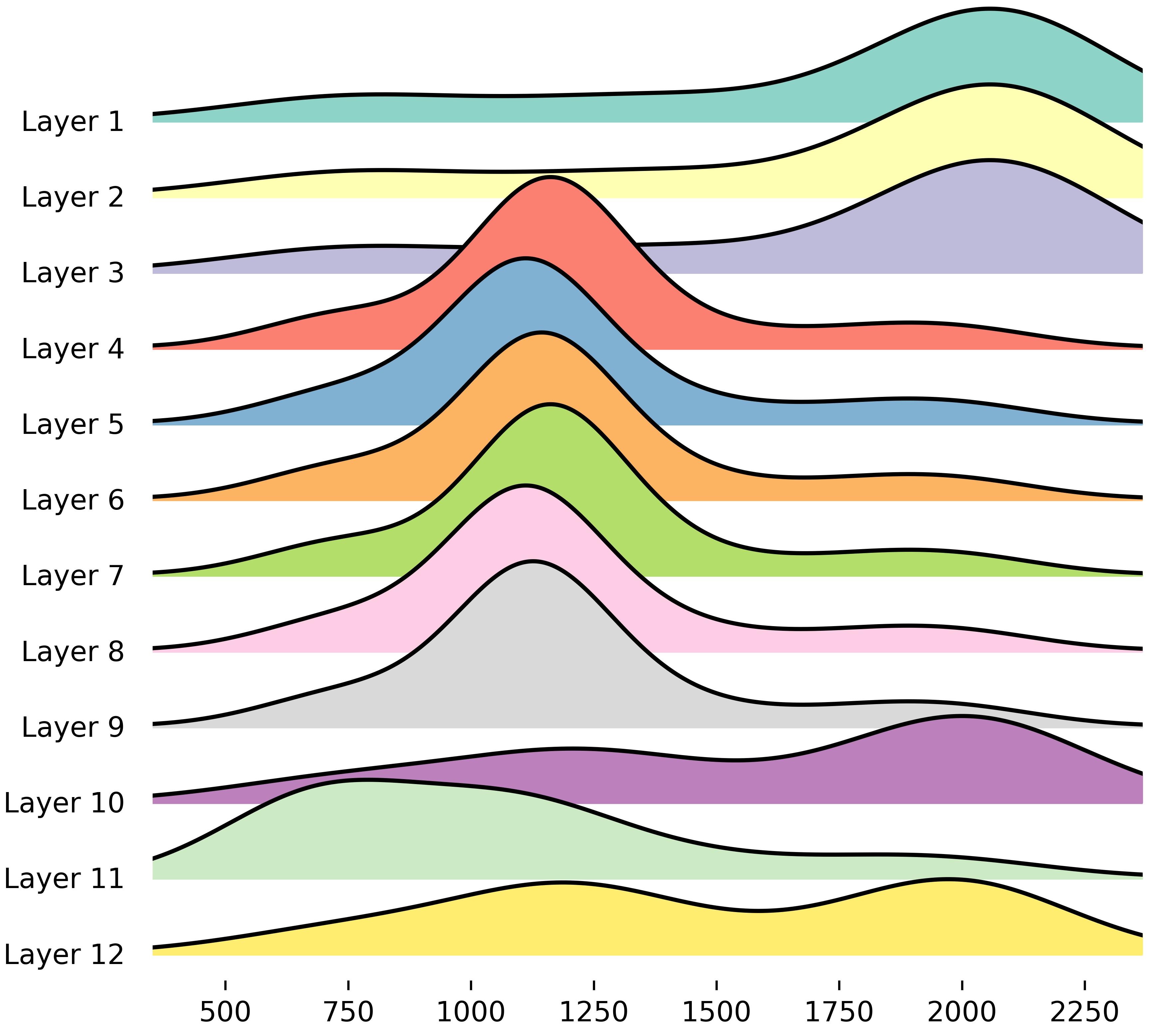}%
    \label{figure2}}
    \hfil
    \subfloat[Dimensional Distribution of Embedding Channels of the Model for the "Church" Category.]
    {\includegraphics[width=1.5in]{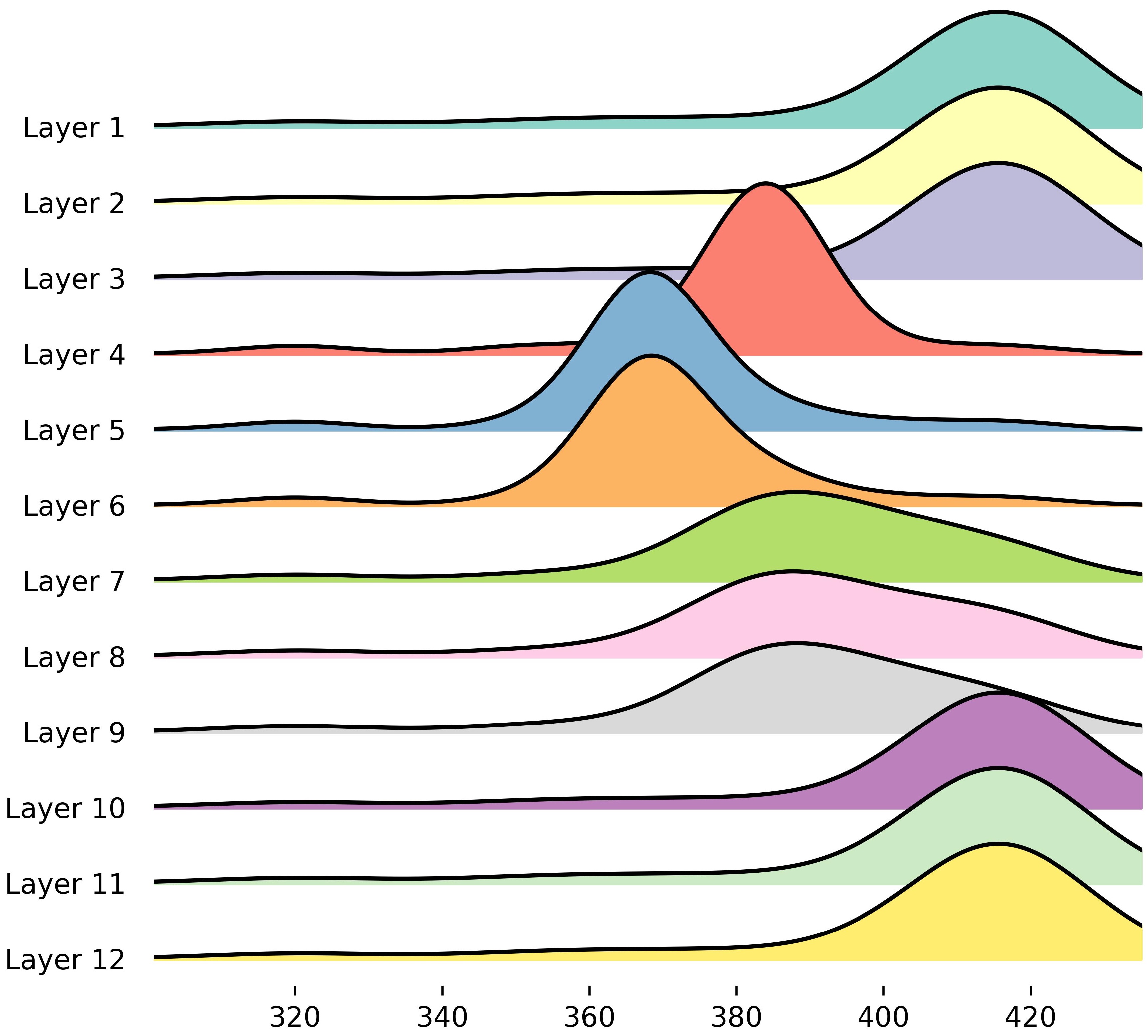}%
    \label{figure3}}
    \hfil
    \subfloat[Dimensional Distribution of MLP of the Model for the "Church" Category.]{\includegraphics[width=1.5in]{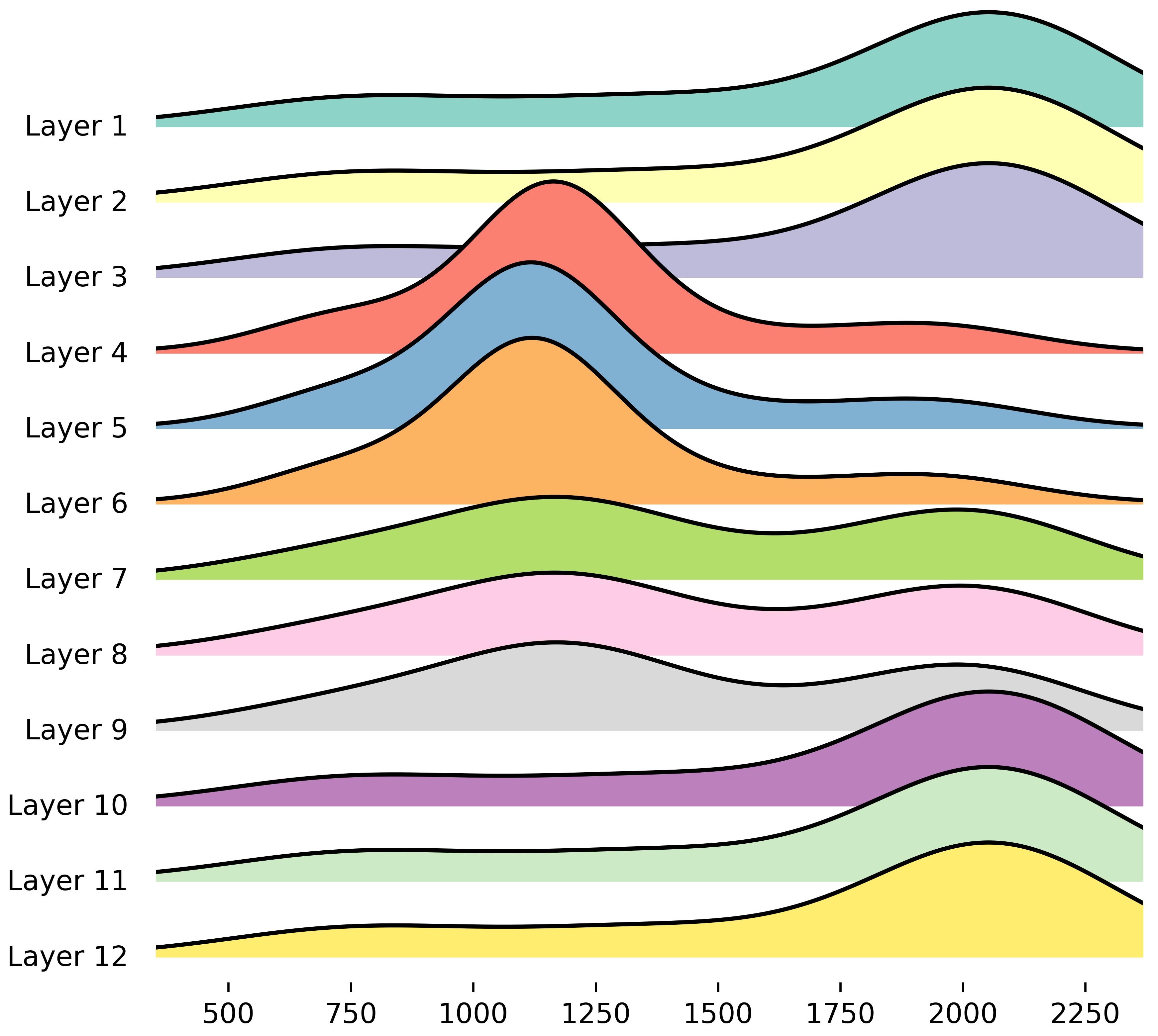}%
    \label{figure4}}
    \caption{
    \textbf{Dimensional distributions of different categories on ImageNet.} 
    The category ``Church'' contains complex architectural information with high information, while the category ``Bird'' is a photograph with a simple background and less information. 
    One can see that \sysname~can dynamically select model channels based on the complexity of the image, achieving the highest possible Top-1 accuracy with the lowest possible FLOPs. 
    }
    \label{Channel distribution}
\end{figure*}

\noindent \textbf{Impact of penalization on token optimization strategies.} An interesting phenomenon was observed during training the \sysname: without penalizing the token optimization terms, the selector tends to retain all tokens, and obtain a high-precision lightweight model by optimizing the model structure. 
The results of the comparative experiments are shown in Tab.~\ref{token_penalty}. 
As the model scale increases, the selector tends to keep more tokens.
In the Tiny scale, a good balance can be achieved even without adding the penalty factor $a_t$.
However, in the Small and Base scales, the proportion of retained tokens is around $80\%$ or higher, sometimes even reaching $100\%$.
The reason may be that adjusting the structure of the meta-network has a smaller impact on accuracy compared to optimizing tokens.
Besides, this phenomenon becomes more pronounced as the model scale increases, which demonstrates that the model has learned effective information from the data. 
As the scale increases, the less information is required from the data. 

Unfortunately, merely optimizing the model structure while retaining a large number of tokens results in a suboptimal state. Although the Top-1 accuracy of the model is maintained, the computational complexity remains high.
By adding penalty terms to the reward function, \sysname~effectively achieves a balance between the number of tokens and the model structure, reducing FLOPs to a lower level while maintaining Top-1 accuracy.

\begin{table}[t]
  \centering
  \caption{The effect of penalization on token optimization.}
  \begin{threeparttable}          
  \scalebox{0.83}{
    \begin{tabular}{cccccc}
    \toprule
    \textbf{Model} & \textbf{Reward State} & \textbf{Type} & \textbf{Top-1 Acc. (\%)} & \textbf{FLOPs(G)} & \textbf{\makecell[c]{Token Keep \\ Rate}} \\
    \midrule
    \multirow{6}[4]{*}{\textbf{Tiny}} & \multirow{3}[2]{*}{Penalty} & Pruning & 66.39\% & 0.99  & 38\% \\
          &       & Merging & 68.85\% & 1.18  & 32\% \\
          &       & P + M & 71.59\% & 1.18  & 60\% \\
\cmidrule{2-6}          & \multirow{3}[2]{*}{Raw} & Pruning & 71.92\% & 1.18  & 60\% \\
          &       & Merging & 70.51\% & 0.92  & 53\% \\
          &       & P + M & 71.15\% & 0.89  & 46\% \\
    \midrule
    \multirow{6}[4]{*}{\textbf{Small}} & \multirow{3}[2]{*}{Penalty} & Pruning & 74.10\% & 2.72  & 13\% \\
          &       & Merging & 76.97\% & 1.98  & 9\% \\
          &       & P + M & 75.28\% & 2.36  & 29\% \\
\cmidrule{2-6}          & \multirow{3}[2]{*}{Raw} & Pruning & 77.57\% & 3.78      & 96\% \\
          &       & Merging & 78.44\% & 4.17  & 97\% \\
          &       & P + M &  76.92\%  & 3.92  & 73\%  \\
    \midrule
    \multirow{6}[4]{*}{\textbf{Base}} & \multirow{3}[2]{*}{Penalty} & Pruning & 78.33\% & 9.18  & 59\% \\
          &       & Merging & 74.61\% & 8.31  & 51\% \\
          &       & P + M & 79.99\% & 10.7  & 30\% \\
\cmidrule{2-6}          & \multirow{3}[2]{*}{Raw} & Pruning & 81.40\%  & 15.18 & 100\% \\
          &       & Merging & 79.90\%  & 13.79  & 88\%  \\
          &       & P + M & 80.90\%    & 14.18  & 93\% \\
    \bottomrule
    \end{tabular}%
    }
    \begin{tablenotes}
        \begin{minipage}[t]{0.46\textwidth}
            \footnotesize              
            \item The impact of token penalties on various ViT models and token optimization strategies. Due to the significant impact of optimizing tokens on accuracy, the selector tends to optimize the model structure and retain tokens. This situation is significantly mitigated after adding token penalties. 
        \end{minipage}
        \end{tablenotes}            
    \end{threeparttable}
  \label{token_penalty}%
\end{table}%

\noindent \textbf{The effect of different features on Selector.} 
Inspired by existing lightweighting efforts, we believe that the intermediate features of the trained ViT can effectively reflect data characteristics. 
Therefore, we attempted to optimize the model structure and tokens on a sample-wise basis, separately utilizing the Class Token and Q, K, and V feature matrices in each ViT block. 
The experiments are conducted on three models of different scales, and the results are presented in Tab. \ref{ppo_input}. 
Due to the maximum embedding dimensions of 240, 416, and 768 for the three model scales, which result in excessive training costs and suboptimal accuracy, this approach was not adopted.
The Q, K, V matrices perform well on different model scales, with the K matrix showing the most significant effect. 
\begin{table}[t]
  \centering
  \caption{The effect of different features on Selector.}
  \begin{threeparttable}          
  \scalebox{0.96}{
    \begin{tabular}{ccccc}
    \toprule
    \textbf{Model} & \textbf{Input} & \textbf{Top-1 Acc. (\%)} & \textbf{FLOPs(G)} & \textbf{Token Keep Rate} \\
    \midrule
    \multirow{3}[2]{*}{Tiny} & Q     & 72.01\% & 0.91  & 42\% \\
          & K     & 72.38\% & 0.87  & 25\% \\
          & V     & 70.09\% & 0.95  & 53\% \\
    \midrule
    \multirow{3}[2]{*}{Small} & Q     & 79.63\% & 2.72  & 30\% \\
          & K     & 80.12\% & 2.84  & 36\% \\
          & V     & 77.97\% & 3.11  & 32\% \\
    \midrule
    \multirow{3}[2]{*}{Base} & Q     & 81.32\% & 10.59  & 64\% \\
          & K     & 81.29\% & 9.60  & 51\% \\
          & V     & 80.90\% & 11.36  & 89\% \\
    \bottomrule
    \end{tabular}%
    }
    \begin{tablenotes}
        \begin{minipage}[t]{0.46\textwidth}
            \footnotesize              
            \item The results of different features as inputs for the selector on ViT-Tiny, Small, and Base models. The $\tt{<CLS>}$ token is discarded due to its high dimensionality and excessive computational overhead. The K matrix performs relatively the best across models of different scales.
        \end{minipage}
        \end{tablenotes}            
    \end{threeparttable}
  \label{ppo_input}%
\end{table}%

\noindent \textbf{The impact of different features on token merging.} 
To improve computational efficiency, we draw on token merging methods from ToMe \cite{tome} and Diffrate \cite{diffrate}, exploring the effects of feature matrices $K$ and $S$ in token importance ranking and merging.
The results are shown in Tab.~\ref{tab: token merging matrices}.
To intuitively demonstrate the impact of the two evaluation metrics on token optimization, the results before fine-tuning are presented.
Overall, matrix $X$ has a more favorable impact on the results compared to matrix $K$, and this effect becomes increasingly pronounced as the model size grows. 
Besides, for both token merging and token pruning-then-merging, the impact of the two matrices is similar.
A possible reason is that, compared to matrix $K$, matrix $X$ not only includes information about token optimization, but also provides more intuitive information about model channel optimization.
From the training perspective, obtaining a high-performance PPO selector based on the $X$ matrix is much easier than using the $K$ matrix, which often requires a very challenging parameter tuning process.

\begin{table}[t]
  \centering
  \caption{The impact of different features on token merging.}
  \begin{threeparttable} 
  \scalebox{0.9}{
    \begin{tabular}{cccccc}
    \toprule
    \textbf{Model} & \textbf{Input} & \textbf{Type} & \textbf{Top-1 Acc. (\%)} & \textbf{FLOPs(G)} & \textbf{\makecell[c]{Token Keep \\ Ratio}} \\
    \midrule
    \multirow{4}[4]{*}{\textbf{Tiny}} & \multirow{2}[2]{*}{K} & Merging & 71.90\% & 0.92  & 62\% \\
          &       & P + M & 73.58\% & 0.99  & 40\% \\
\cmidrule{2-6}          & \multirow{2}[2]{*}{X} & Merging & 72.81\% & 0.96  & 53\% \\
          &       & P + M & 74.36\% & 1.00  & 33\% \\
    \midrule
    \multirow{4}[4]{*}{\textbf{Small}} & \multirow{2}[2]{*}{K} & Merging & 78.72\% & 2.64  & 28\% \\
          &       & P + M & 78.82\% & 2.90  & 44\% \\
\cmidrule{2-6}          & \multirow{2}[2]{*}{X} & Merging & 80.17\% & 2.85  & 38\% \\
          &       & P + M & 80.19\% & 2.96  & 33\% \\
    \midrule
    \multirow{4}[4]{*}{\textbf{Base}} & \multirow{2}[2]{*}{K} & Merging & 79.70\% & 9.12  & 54\% \\
          &       & P + M & 79.37\% & 8.22  & 56\% \\
\cmidrule{2-6}          & \multirow{2}[2]{*}{X} & Merging & 81.57\% & 9.64  & 65\% \\
          &       & P + M & 77.02\%  & 11.85  & 74\%  \\
    \bottomrule
    \end{tabular}%
    }
    \begin{tablenotes}
        \begin{minipage}[t]{0.46\textwidth}
            \footnotesize              
            \item The impact of various selector inputs on token merging and pruning-then-merging across ViT models (before fine-tuning). The performance of adopting $X$ is significantly better than that of matrix $K$, and this difference becomes more pronounced as the model size increases. 
        \end{minipage}
        \end{tablenotes}            
    \end{threeparttable}
  \label{tab: token merging matrices}%
\end{table}%

\noindent \textbf{The effect of smooth accuracy.} 
The Top-1 accuracy is modified by Eq.~(\ref{loss function}) to avoid the influence of meta-network performance. 
Tab.~\ref{acc_effect} shows the comparison results. 
Note that these results are obtained before fine-tuning, which can more clearly demonstrate the impact of the smoothing function.
Overall, the performance of using the smoothing function is superior to not using it, and this effect becomes increasingly pronounced as the model scale increases. 
At the Tiny scale, smooth function has no significant impact. 
While at the Small scale, the impact is only evident in the pruning strategy, causing both the pruning and pruning-then-merging strategies to maintain high Top-1 accuracy, but at the cost of increased FLOPs. Ultimately, at the Base scale, this impact extends to all token optimization strategies.

\begin{table}[t]
  \centering
  \caption{The effect of smooth accuracy.}
  \begin{threeparttable}          
  \scalebox{0.83}{
    \begin{tabular}{cccccc}
    \toprule
    \textbf{Model} & \textbf{Reward State} & \textbf{Type} & \textbf{Top-1 Acc. (\%)} & \textbf{FLOPs(G)} & \textbf{\makecell[c]{Token Keep \\ Rate}} \\
    \midrule
    \multirow{6}[4]{*}{\textbf{Tiny}} & \multirow{3}[2]{*}{Raw} & Pruning & 67.67\% & 1.01  & 59\% \\
          &       & Merging & 71.91\% & 0.96  & 26\% \\
          &       & P + M & 73.40\% & 1.18  & 31\% \\
\cmidrule{2-6}          & \multirow{3}[2]{*}{Smooth} & Pruning & 70.62\% & 1.07  & 60\% \\
          &       & Merging & 71.07\% & 0.92  & 35\% \\
          &       & P + M & 72.67\% & 1.01  & 42\% \\
    \midrule
    \multirow{6}[4]{*}{\textbf{Small}} & \multirow{3}[2]{*}{Raw} & Pruning & 72.99\% & 3.26  & 76\% \\
          &       & Merging & 75.38\% & 2.15  & 35\% \\
          &       & P + M & 68.74\% & 2.97  & 56\% \\
\cmidrule{2-6}          & \multirow{3}[2]{*}{Smooth} & Pruning & 74.10\% & 2.70  & 13\% \\
          &       & Merging & 75.48\% & 2.07  & 24\% \\
          &       & P + M & 75.28\% & 2.36  & 29\% \\
    \midrule
    \multirow{6}[4]{*}{\textbf{Base}} & \multirow{3}[2]{*}{Raw} & Pruning & 81.04\% & 14.50  & 55\% \\
          &       & Merging & 79.46\% & 12.65  & 75\% \\
          &       & P + M & 79.62\% & 12.52  & 94\% \\
\cmidrule{2-6}          & \multirow{3}[2]{*}{Smooth} & Pruning & 81.08\% & 11.50  & 51\% \\
          &       & Merging & 79.22\% & 10.45  & 71\% \\
          &       & P + M & 79.99\% & 10.69  & 30\% \\
    \bottomrule
    \end{tabular}%
    }
    \begin{tablenotes}
        \begin{minipage}[t]{0.46\textwidth}
            \footnotesize              
            \item The impact of the smoothing function on model performance across various ViT models and various token optimization strategies before fine-tuning. The smoothing function has little impact on the Tiny model but significantly affects the Small and Base models. On the other hand, its impact is more pronounced on the pruning strategy compared to the merging strategy.
        \end{minipage}
        \end{tablenotes}            
    \end{threeparttable}
  \label{acc_effect}%
\end{table}%

\begin{figure*}[t]
    \centering
    \subfloat[ViT-Tiny]
    {\includegraphics[width=2.2in]{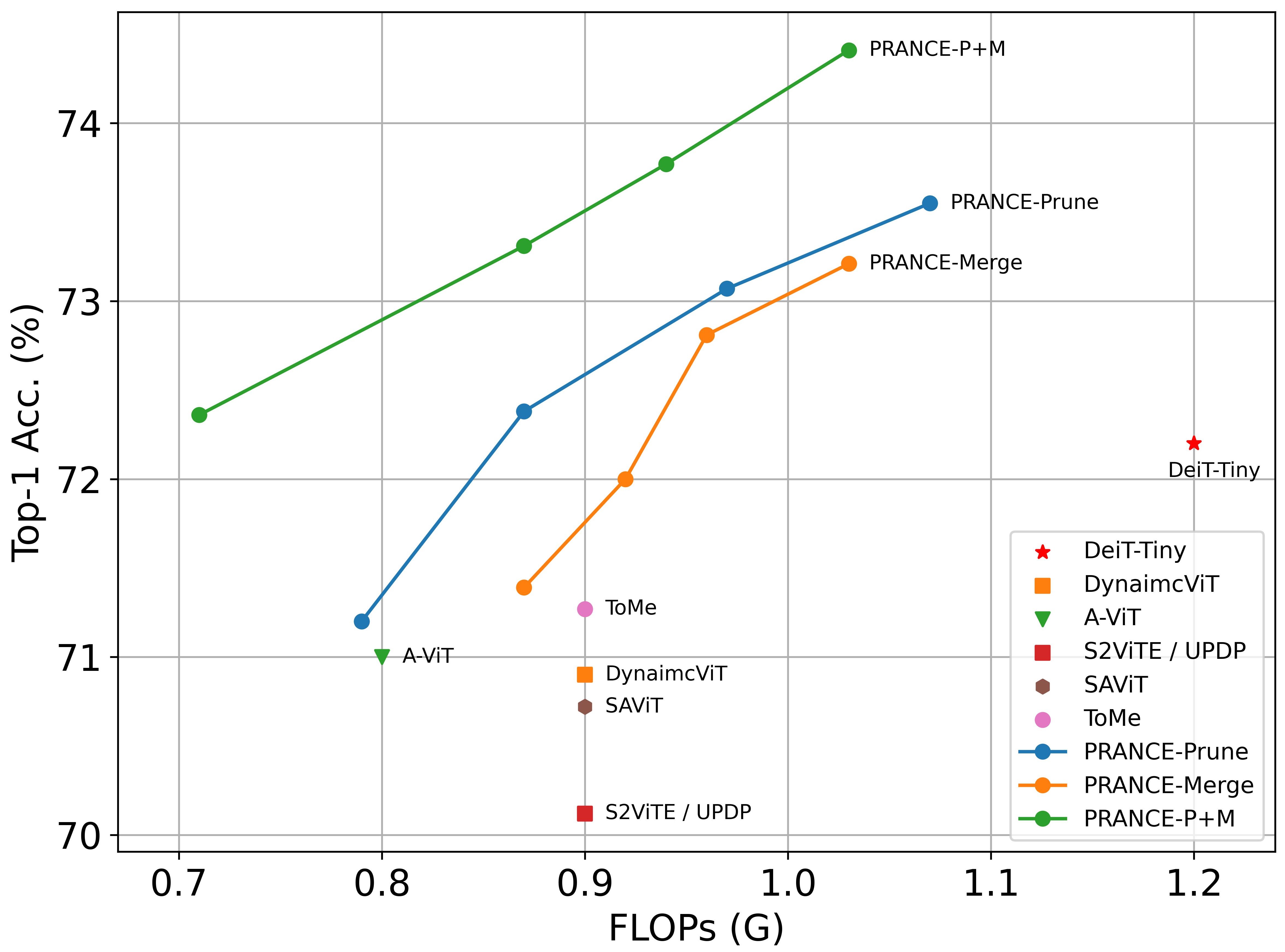}%
    }
    \hfil
    \subfloat[ViT-Small]
    {\includegraphics[width=2.2in]{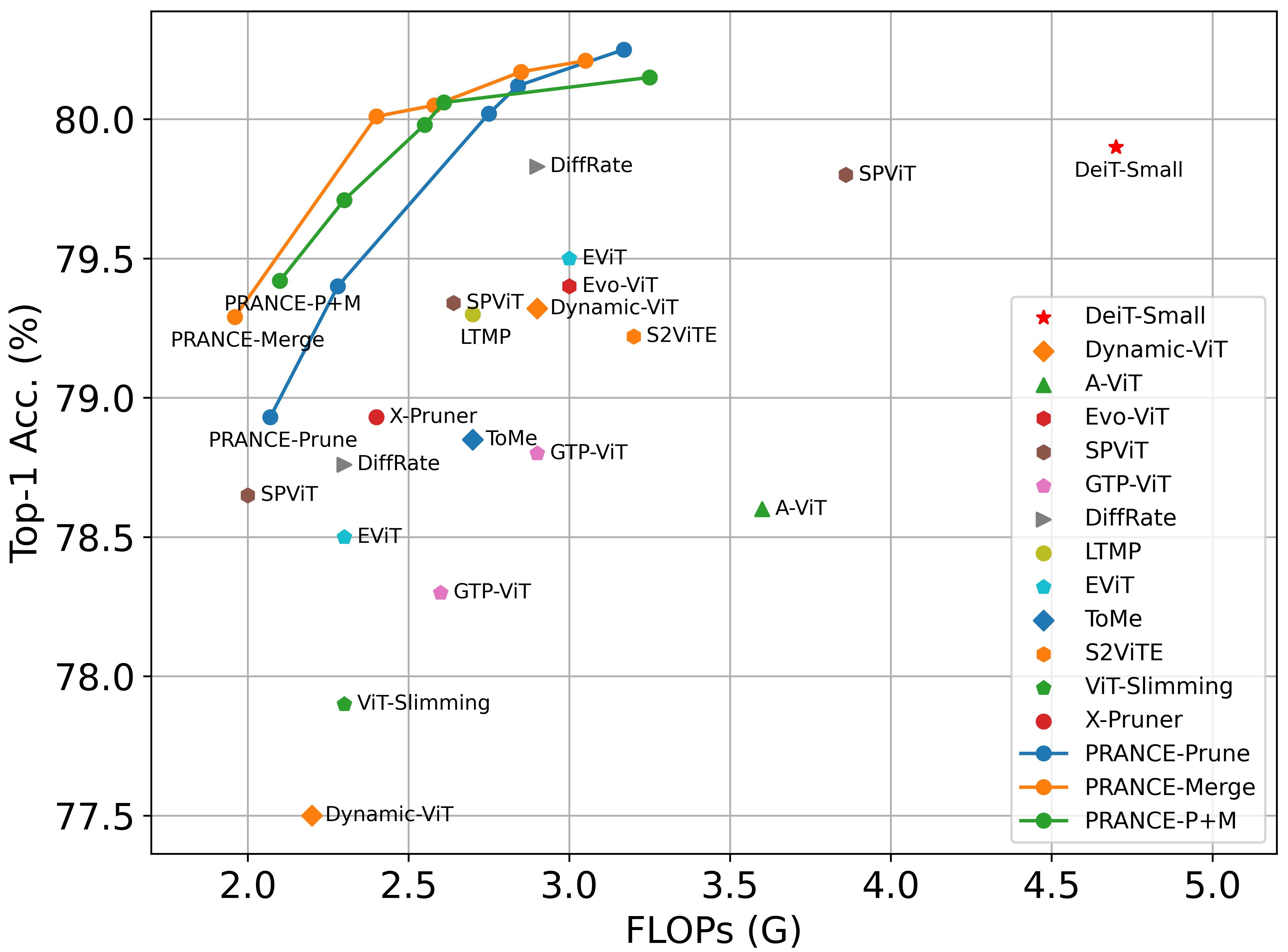}%
    }
    \hfil
    \subfloat[ViT-Base]
    {\includegraphics[width=2.25in]{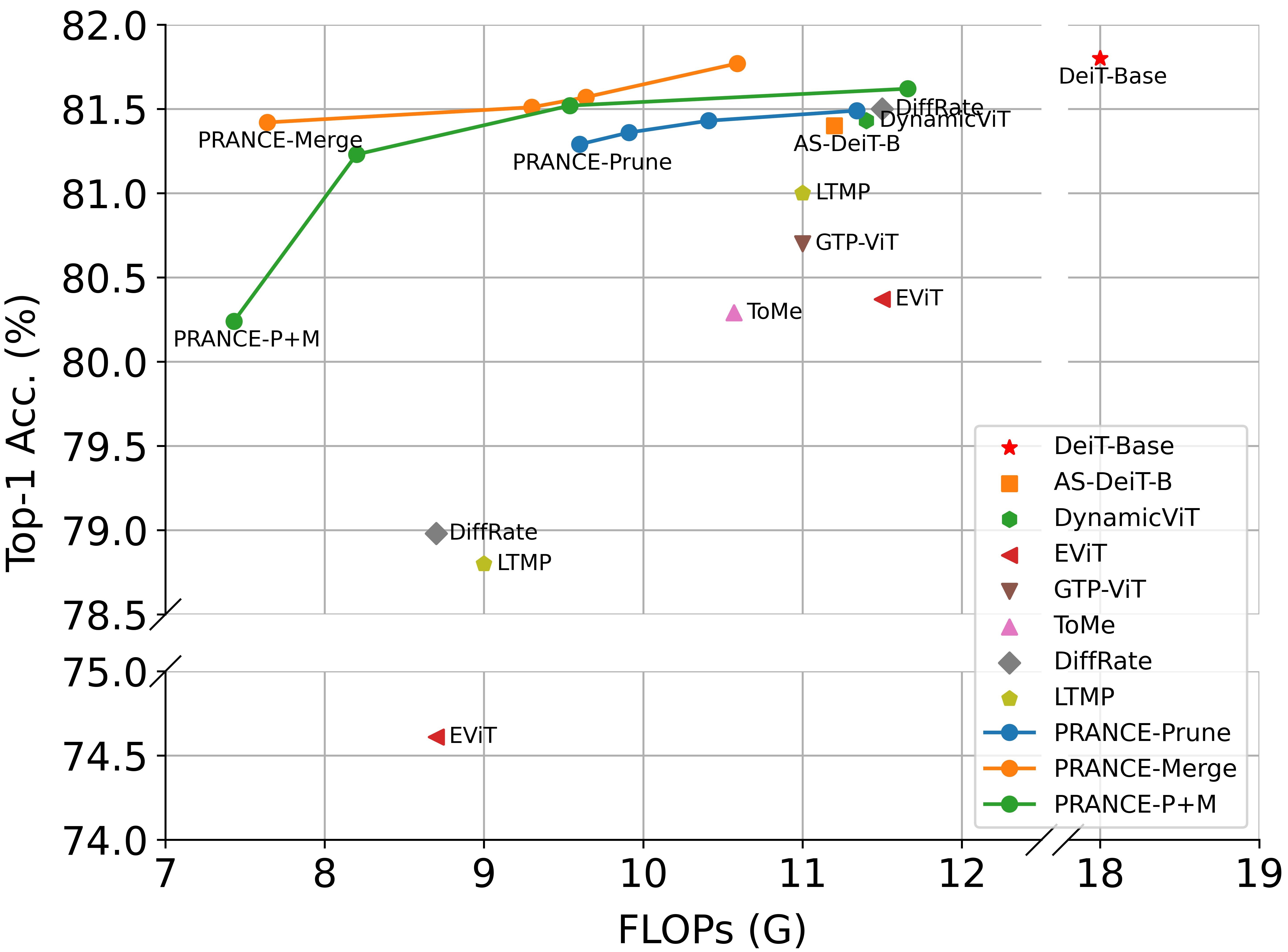}%
    }
    \caption{
    \textbf{Accuracy-FLOPs performance.}
    Across different model scales, \sysname~achieves higher Top-1 accuracy with lower FLOPs, surpassing various SOTA methods and even exceeding DeiT at the Tiny and Small scales. On the other hand, as the model scale increases, the margin by which \sysname~surpasses DeiT becomes smaller. This indicates that the model is learning more, and the amount of data becomes insufficient at the Base scale.
    }
    \label{acc_FLOPs_chart}
\end{figure*}

\noindent \textbf{Visualization.} In this part, some visualization results will be shown. 
First of all, Fig.~\ref{pruning_vis}, Fig.~\ref{merging_vis}, Fig.~\ref{pruning_merging_vis} present the step-by-step visualization results of \sysname~on different samples using the pruning, merging, and pruning-then-merging strategies, respectively. 
All three strategies can dynamically optimize different numbers of tokens based on the complexity of the sample at various stages of model inference, retaining important tokens and removing unimportant ones. 
While there are still some differences between them. 
The pruning strategy primarily optimizes tokens in the later stages, specifically at the 6th and 9th blocks, while retaining more tokens at the 3rd block.
The pruning-then-merging strategy, on the other hand, tends to optimize tokens in the early stages, with the lightest optimization at the 9th layer.
In contrast, the merging strategy has a more balanced optimization process.
This phenomenon is closely related to the optimization of the model structure. When the model has acquired sufficient information, it tends to optimize more tokens. Conversely, the model prefer to use as many tokens as possible to capture sample information.

Fig.~\ref{Channel distribution} shows the distribution of channels in the embedding layers and MLP layers. 
To ensure a fair comparison, we fixed the first three layers to their maximum structure, while the remaining nine layers were optimized. 
Samples in the "Bird" category are relatively simple: the main subject of the image is complex, but the background is very simple. 
In contrast, samples in the "Church" category are more complex, with the entire image filled with intricate architectural details.
It can be observed that for complex samples, \sysname~tends to use more channels, while for simple samples, the number of activated channels is significantly reduced. 
It demonstrates that \sysname~can dynamically adjust the model's complexity based on different samples.
What's more, the dynamic adjustment of the model structure by \sysname~is more evident in samples of varying complexity rather than in samples of different categories. This is because samples from different categories can contain both simple and complex images. 

Fig.~\ref{acc_FLOPs_chart} illustrates the correlation curves between FLOPs and accuracy under three different scales. 
The models closer to the top-left corner in the figure exhibit better performance.
Overall, \sysname~achieves excellent results across different model scales by employing various token optimization methods. 
It not only surpasses the vast majority of SOTA lightweight algorithms but also matches or exceeds the performance of the baseline model DeiT with significantly lower FLOPs.
On the other hand, the different token optimization methods can all achieve good results, with no significant differences in their effectiveness. 
This is related to the selector’s collaborative optimization of the model structure and tokens, as well as the complementary information in different token optimization modes.
\section{Conclusion}
In this paper, we propose the \sysname~framework for ViT compression that optimizes both the architecture (model channels) and data (number of tokens). 
To this end, we pre-train a meta-network that supports variable channels and then model the inference process of ViTs as a Markov decision process, using PPO as the selector, and propose a matching “Result-to-Go” training mechanism. 
\sysname~supports the joint optimization of the model structure and three different token optimization methods: pruning, merging, and pruning-merging, all of which yield good results.
Extensive experiments have demonstrated the outstanding performance of this framework the performance of \sysname~not surpasses SOTA methods, demonstrating significant potential for widespread applications. 

\balance
\bibliographystyle{IEEEtran}
\bibliography{IEEEabrv,reference}

\begin{thebibliography}{10}
\providecommand{\url}[1]{#1}
\csname url@samestyle\endcsname
\providecommand{\newblock}{\relax}
\providecommand{\bibinfo}[2]{#2}
\providecommand{\BIBentrySTDinterwordspacing}{\spaceskip=0pt\relax}
\providecommand{\BIBentryALTinterwordstretchfactor}{4}
\providecommand{\BIBentryALTinterwordspacing}{\spaceskip=\fontdimen2\font plus
\BIBentryALTinterwordstretchfactor\fontdimen3\font minus \fontdimen4\font\relax}
\providecommand{\BIBforeignlanguage}[2]{{%
\expandafter\ifx\csname l@#1\endcsname\relax
\typeout{** WARNING: IEEEtran.bst: No hyphenation pattern has been}%
\typeout{** loaded for the language `#1'. Using the pattern for}%
\typeout{** the default language instead.}%
\else
\language=\csname l@#1\endcsname
\fi
#2}}
\providecommand{\BIBdecl}{\relax}
\BIBdecl

\bibitem{han2022survey}
K.~Han, Y.~Wang, H.~Chen, X.~Chen, J.~Guo, Z.~Liu, Y.~Tang, A.~Xiao, C.~Xu, Y.~Xu \emph{et~al.}, ``A survey on vision transformer,'' \emph{IEEE Transactions on Pattern Analysis and Machine Intelligence}, vol.~45, no.~1, pp. 87--110, 2022.

\bibitem{vitclass}
A.~Dosovitskiy, L.~Beyer, A.~Kolesnikov, D.~Weissenborn, X.~Zhai, T.~Unterthiner, M.~Dehghani, M.~Minderer, G.~Heigold, S.~Gelly, J.~Uszkoreit, and N.~Houlsby, ``An image is worth 16x16 words: Transformers for image recognition at scale,'' in \emph{9th International Conference on Learning Representations, {ICLR} 2021, Virtual Event, Austria, May 3-7, 2021}.\hskip 1em plus 0.5em minus 0.4em\relax OpenReview.net, 2021.

\bibitem{vitdetect}
M.~Gehrig and D.~Scaramuzza, ``Recurrent vision transformers for object detection with event cameras,'' in \emph{{IEEE/CVF} Conference on Computer Vision and Pattern Recognition, {CVPR} 2023, Vancouver, BC, Canada, June 17-24, 2023}.\hskip 1em plus 0.5em minus 0.4em\relax {IEEE}, 2023, pp. 13\,884--13\,893.

\bibitem{vit-detection-2}
Z.~Dai, B.~Cai, Y.~Lin, and J.~Chen, ``Unsupervised pre-training for detection transformers,'' \emph{IEEE Transactions on Pattern Analysis and Machine Intelligence}, vol.~45, no.~11, pp. 12\,772--12\,782, 2023.

\bibitem{vitseg}
M.~K.~H. Thisanke, L.~A.~C. Deshan, K.~Chamith, S.~Seneviratne, R.~Vidanaarachchi, and D.~Herath, ``Semantic segmentation using vision transformers: {A} survey,'' \emph{Eng. Appl. Artif. Intell.}, vol. 126, p. 106669, 2023.

\bibitem{seg-2}
H.~Ding, C.~Liu, S.~Wang, and X.~Jiang, ``Vlt: Vision-language transformer and query generation for referring segmentation,'' \emph{IEEE Transactions on Pattern Analysis and Machine Intelligence}, vol.~45, no.~6, pp. 7900--7916, 2023.

\bibitem{vitmm}
Y.~Wang, X.~Chen, L.~Cao, W.~Huang, F.~Sun, and Y.~Wang, ``Multimodal token fusion for vision transformers,'' in \emph{{IEEE/CVF} Conference on Computer Vision and Pattern Recognition, {CVPR} 2022, New Orleans, LA, USA, June 18-24, 2022}.\hskip 1em plus 0.5em minus 0.4em\relax {IEEE}, 2022, pp. 12\,176--12\,185.

\bibitem{vlt-2}
J.~Deng, Z.~Yang, D.~Liu, T.~Chen, W.~Zhou, Y.~Zhang, H.~Li, and W.~Ouyang, ``Transvg++: End-to-end visual grounding with language conditioned vision transformer,'' \emph{IEEE Transactions on Pattern Analysis and Machine Intelligence}, vol.~45, no.~11, pp. 13\,636--13\,652, 2023.

\bibitem{vit-pruning}
M.~Zhu, Y.~Tang, and K.~Han, ``Vision transformer pruning,'' \emph{arXiv preprint arXiv:2104.08500}, 2021.

\bibitem{vit_wd_pruning}
F.~Yu, K.~Huang, M.~Wang, Y.~Cheng, W.~Chu, and L.~Cui, ``Width \& depth pruning for vision transformers,'' in \emph{Proceedings of the AAAI Conference on Artificial Intelligence}, vol.~36, no.~3, 2022, pp. 3143--3151.

\bibitem{ptq-vit}
Z.~Liu, Y.~Wang, K.~Han, W.~Zhang, S.~Ma, and W.~Gao, ``Post-training quantization for vision transformer,'' \emph{Advances in Neural Information Processing Systems}, vol.~34, pp. 28\,092--28\,103, 2021.

\bibitem{tang2022mixed}
C.~Tang, K.~Ouyang, Z.~Wang, Y.~Zhu, W.~Ji, Y.~Wang, and W.~Zhu, ``Mixed-precision neural network quantization via learned layer-wise importance,'' in \emph{European Conference on Computer Vision}.\hskip 1em plus 0.5em minus 0.4em\relax Springer, 2022, pp. 259--275.

\bibitem{tang2024retraining}
C.~Tang, Y.~Meng, J.~Jiang, S.~Xie, R.~Lu, X.~Ma, Z.~Wang, and W.~Zhu, ``Retraining-free model quantization via one-shot weight-coupling learning,'' in \emph{Proceedings of the IEEE/CVF Conference on Computer Vision and Pattern Recognition}, 2024, pp. 15\,855--15\,865.

\bibitem{Autoformer}
M.~Chen, H.~Peng, J.~Fu, and H.~Ling, ``Autoformer: Searching transformers for visual recognition,'' in \emph{Proceedings of the IEEE/CVF international conference on computer vision}, 2021, pp. 12\,270--12\,280.

\bibitem{Elasticvit}
C.~Tang, L.~L. Zhang, H.~Jiang, J.~Xu, T.~Cao, Q.~Zhang, Y.~Yang, Z.~Wang, and M.~Yang, ``Elasticvit: Conflict-aware supernet training for deploying fast vision transformer on diverse mobile devices,'' in \emph{Proceedings of the IEEE/CVF International Conference on Computer Vision}, 2023, pp. 5829--5840.

\bibitem{yu2020bignas}
J.~Yu, P.~Jin, H.~Liu, G.~Bender, P.-J. Kindermans, M.~Tan, T.~Huang, X.~Song, R.~Pang, and Q.~Le, ``Bignas: Scaling up neural architecture search with big single-stage models,'' in \emph{Computer Vision--ECCV 2020: 16th European Conference, Glasgow, UK, August 23--28, 2020, Proceedings, Part VII 16}.\hskip 1em plus 0.5em minus 0.4em\relax Springer, 2020, pp. 702--717.

\bibitem{nvit}
H.~Yang, H.~Yin, P.~Molchanov, H.~Li, and J.~Kautz, ``Nvit: Vision transformer compression and parameter redistribution,'' \emph{CoRR}, vol. abs/2110.04869, 2021.

\bibitem{wang2021attentivenas}
D.~Wang, M.~Li, C.~Gong, and V.~Chandra, ``Attentivenas: Improving neural architecture search via attentive sampling,'' in \emph{Proceedings of the IEEE/CVF conference on computer vision and pattern recognition}, 2021, pp. 6418--6427.

\bibitem{dynamicvit}
Y.~Rao, Z.~Liu, W.~Zhao, J.~Zhou, and J.~Lu, ``Dynamic spatial sparsification for efficient vision transformers and convolutional neural networks,'' \emph{IEEE Transactions on Pattern Analysis and Machine Intelligence}, vol.~45, no.~9, pp. 10\,883--10\,897, 2023.

\bibitem{liang2022evit}
Y.~Liang, C.~Ge, Z.~Tong, Y.~Song, J.~Wang, and P.~Xie, ``Not all patches are what you need: Expediting vision transformers via token reorganizations,'' in \emph{International Conference on Learning Representations}, 2022.

\bibitem{haykin1994neural}
S.~Haykin, \emph{Neural networks: a comprehensive foundation}.\hskip 1em plus 0.5em minus 0.4em\relax Prentice Hall PTR, 1994.

\bibitem{nipsw23}
H.~Wang, J.~Fan, Z.~Chen, H.~Li, W.~Liu, T.~Liu, Q.~Dai, Y.~Wang, Z.~Dong, and R.~Tang, ``Optimal transport for treatment effect estimation,'' in \emph{Advances in Neural Information Processing Systems 36: Annual Conference on Neural Information Processing Systems 2023, NeurIPS 2023, New Orleans, LA, USA, December 10 - 16, 2023}, A.~Oh, T.~Naumann, A.~Globerson, K.~Saenko, M.~Hardt, and S.~Levine, Eds., 2023.

\bibitem{tome}
D.~Bolya, C.-Y. Fu, X.~Dai, P.~Zhang, C.~Feichtenhofer, and J.~Hoffman, ``Token merging: Your vit but faster,'' \emph{arXiv preprint arXiv:2210.09461}, 2022.

\bibitem{bolya2023token}
D.~Bolya and J.~Hoffman, ``Token merging for fast stable diffusion,'' in \emph{Proceedings of the IEEE/CVF Conference on Computer Vision and Pattern Recognition}, 2023, pp. 4598--4602.

\bibitem{diffrate}
M.~Chen, W.~Shao, P.~Xu, M.~Lin, K.~Zhang, F.~Chao, R.~Ji, Y.~Qiao, and P.~Luo, ``Diffrate: Differentiable compression rate for efficient vision transformers,'' in \emph{Proceedings of the IEEE/CVF International Conference on Computer Vision}, 2023, pp. 17\,164--17\,174.

\bibitem{long2023beyond}
S.~Long, Z.~Zhao, J.~Pi, S.~Wang, and J.~Wang, ``Beyond attentive tokens: Incorporating token importance and diversity for efficient vision transformers,'' in \emph{Proceedings of the IEEE/CVF Conference on Computer Vision and Pattern Recognition}, 2023, pp. 10\,334--10\,343.

\bibitem{wang2023zero}
H.~Wang, B.~Dedhia, and N.~K. Jha, ``Zero-tprune: Zero-shot token pruning through leveraging of the attention graph in pre-trained transformers,'' \emph{arXiv preprint arXiv:2305.17328}, 2023.

\bibitem{vit-slimming}
A.~Chavan, Z.~Shen, Z.~Liu, Z.~Liu, K.-T. Cheng, and E.~P. Xing, ``Vision transformer slimming: Multi-dimension searching in continuous optimization space,'' in \emph{Proceedings of the IEEE/CVF Conference on Computer Vision and Pattern Recognition}, 2022, pp. 4931--4941.

\bibitem{fq-vit}
Y.~Lin, T.~Zhang, P.~Sun, Z.~Li, and S.~Zhou, ``Fq-vit: Post-training quantization for fully quantized vision transformer,'' \emph{arXiv preprint arXiv:2111.13824}, 2021.

\bibitem{tang2022arbitrary}
C.~Tang, H.~Zhai, K.~Ouyang, Z.~Wang, Y.~Zhu, and W.~Zhu, ``Arbitrary bit-width network: A joint layer-wise quantization and adaptive inference approach,'' in \emph{Proceedings of the 30th ACM International Conference on Multimedia}, 2022, pp. 2899--2908.

\bibitem{Efficientvit}
X.~Liu, H.~Peng, N.~Zheng, Y.~Yang, H.~Hu, and Y.~Yuan, ``Efficientvit: Memory efficient vision transformer with cascaded group attention,'' in \emph{Proceedings of the IEEE/CVF Conference on Computer Vision and Pattern Recognition}, 2023, pp. 14\,420--14\,430.

\bibitem{token_pruning_initial}
Y.~Liang, C.~Ge, Z.~Tong, Y.~Song, J.~Wang, and P.~Xie, ``Not all patches are what you need: Expediting vision transformers via token reorganizations,'' \emph{arXiv preprint arXiv:2202.07800}, 2022.

\bibitem{A-vit}
H.~Yin, A.~Vahdat, J.~M. Alvarez, A.~Mallya, J.~Kautz, and P.~Molchanov, ``A-vit: Adaptive tokens for efficient vision transformer,'' in \emph{Proceedings of the IEEE/CVF Conference on Computer Vision and Pattern Recognition}, 2022, pp. 10\,809--10\,818.

\bibitem{Evo_vit}
Y.~Xu, Z.~Zhang, M.~Zhang, K.~Sheng, K.~Li, W.~Dong, L.~Zhang, C.~Xu, and X.~Sun, ``Evo-vit: Slow-fast token evolution for dynamic vision transformer,'' in \emph{Proceedings of the AAAI Conference on Artificial Intelligence}, vol.~36, no.~3, 2022, pp. 2964--2972.

\bibitem{multi_scale_token}
J.~B. Haurum, M.~Madadi, S.~Escalera, and T.~B. Moeslund, ``Multi-scale hybrid vision transformer and sinkhorn tokenizer for sewer defect classification,'' \emph{Automation in Construction}, vol. 144, p. 104614, 2022.

\bibitem{token_pooling}
D.~Marin, J.-H.~R. Chang, A.~Ranjan, A.~Prabhu, M.~Rastegari, and O.~Tuzel, ``Token pooling in vision transformers for image classification,'' in \emph{Proceedings of the IEEE/CVF Winter Conference on Applications of Computer Vision}, 2023, pp. 12--21.

\bibitem{merge_token_vit}
C.~Renggli, A.~S. Pinto, N.~Houlsby, B.~Mustafa, J.~Puigcerver, and C.~Riquelme, ``Learning to merge tokens in vision transformers,'' \emph{arXiv preprint arXiv:2202.12015}, 2022.

\bibitem{vaswani2017attention}
A.~Vaswani, N.~Shazeer, N.~Parmar, J.~Uszkoreit, L.~Jones, A.~N. Gomez, {\L}.~Kaiser, and I.~Polosukhin, ``Attention is all you need,'' \emph{Advances in neural information processing systems}, vol.~30, 2017.

\bibitem{lbc}
J.~Fan, Y.~Zhuang, Y.~Liu, J.~Hao, B.~Wang, J.~Zhu, H.~Wang, and S.~Xia, ``Learnable behavior control: Breaking atari human world records via sample-efficient behavior selection,'' in \emph{The Eleventh International Conference on Learning Representations, {ICLR} 2023, Kigali, Rwanda, May 1-5, 2023}.\hskip 1em plus 0.5em minus 0.4em\relax OpenReview.net, 2023.

\bibitem{reviewatari}
J.~Fan, ``A review for deep reinforcement learning in atari: Benchmarks, challenges, and solutions,'' \emph{CoRR}, vol. abs/2112.04145, 2021.

\bibitem{lee_ddpg}
Y.~Li, Z.~Liu, G.~Lan, M.~Sader, and Z.~Chen, ``A ddpg-based solution for optimal consensus of continuous-time linear multi-agent systems,'' \emph{Science China Technological Sciences}, vol.~66, no.~8, pp. 2441--2453, 2023.

\bibitem{lee_td3}
Z.~Liu, Y.~Li, G.~Lan, and Z.~Chen, ``A novel data-driven model-free synchronization protocol for discrete-time multi-agent systems via td3 based algorithm,'' \emph{Knowledge-Based Systems}, vol. 287, p. 111430, 2024.

\bibitem{asp_solver}
C.~Wang, Z.~Yu, S.~McAleer, T.~Yu, and Y.~Yang, ``Asp: Learn a universal neural solver!'' \emph{IEEE Transactions on Pattern Analysis and Machine Intelligence}, vol.~46, no.~6, pp. 4102--4114, 2024.

\bibitem{gdi}
J.~Fan and C.~Xiao, ``Generalized data distribution iteration,'' in \emph{International Conference on Machine Learning, {ICML} 2022, 17-23 July 2022, Baltimore, Maryland, {USA}}, ser. Proceedings of Machine Learning Research, K.~Chaudhuri, S.~Jegelka, L.~Song, C.~Szepesv{\'{a}}ri, G.~Niu, and S.~Sabato, Eds., vol. 162.\hskip 1em plus 0.5em minus 0.4em\relax {PMLR}, 2022, pp. 6103--6184.

\bibitem{ppo}
J.~Schulman, F.~Wolski, P.~Dhariwal, A.~Radford, and O.~Klimov, ``Proximal policy optimization algorithms,'' \emph{arXiv preprint arXiv:1707.06347}, 2017.

\bibitem{rlhf}
L.~Ouyang, J.~Wu, X.~Jiang, D.~Almeida, C.~L. Wainwright, P.~Mishkin, C.~Zhang, S.~Agarwal, K.~Slama, A.~Ray, J.~Schulman, J.~Hilton, F.~Kelton, L.~Miller, M.~Simens, A.~Askell, P.~Welinder, P.~F. Christiano, J.~Leike, and R.~Lowe, ``Training language models to follow instructions with human feedback,'' in \emph{Advances in Neural Information Processing Systems 35: Annual Conference on Neural Information Processing Systems 2022, NeurIPS 2022, New Orleans, LA, USA, November 28 - December 9, 2022}, S.~Koyejo, S.~Mohamed, A.~Agarwal, D.~Belgrave, K.~Cho, and A.~Oh, Eds., 2022.

\bibitem{GAE}
J.~Schulman, P.~Moritz, S.~Levine, M.~Jordan, and P.~Abbeel, ``High-dimensional continuous control using generalized advantage estimation,'' \emph{arXiv preprint arXiv:1506.02438}, 2015.

\bibitem{Advantage_Normalization}
G.~Tucker, S.~Bhupatiraju, S.~Gu, R.~Turner, Z.~Ghahramani, and S.~Levine, ``The mirage of action-dependent baselines in reinforcement learning,'' in \emph{International conference on machine learning}.\hskip 1em plus 0.5em minus 0.4em\relax PMLR, 2018, pp. 5015--5024.

\bibitem{rethink}
J.~Fan, C.~Xiao, and Y.~Huang, ``{GDI:} rethinking what makes reinforcement learning different from supervised learning,'' \emph{CoRR}, vol. abs/2106.06232, 2021.

\bibitem{decision_transformer}
L.~Chen, K.~Lu, A.~Rajeswaran, K.~Lee, A.~Grover, M.~Laskin, P.~Abbeel, A.~Srinivas, and I.~Mordatch, ``Decision transformer: Reinforcement learning via sequence modeling,'' \emph{Advances in neural information processing systems}, vol.~34, pp. 15\,084--15\,097, 2021.

\bibitem{contextual_transformer}
R.~Lin, Y.~Li, X.~Feng, Z.~Zhang, X.~H.~W. Fung, H.~Zhang, J.~Wang, Y.~Du, and Y.~Yang, ``Contextual transformer for offline meta reinforcement learning,'' \emph{arXiv preprint arXiv:2211.08016}, 2022.

\bibitem{touvron2021training}
H.~Touvron, M.~Cord, M.~Douze, F.~Massa, A.~Sablayrolles, and H.~J{\'e}gou, ``Training data-efficient image transformers \& distillation through attention,'' in \emph{International conference on machine learning}.\hskip 1em plus 0.5em minus 0.4em\relax PMLR, 2021, pp. 10\,347--10\,357.

\bibitem{savit}
C.~Zheng, K.~Zhang, Z.~Yang, W.~Tan, J.~Xiao, Y.~Ren, S.~Pu \emph{et~al.}, ``Savit: Structure-aware vision transformer pruning via collaborative optimization,'' \emph{Advances in Neural Information Processing Systems}, vol.~35, pp. 9010--9023, 2022.

\bibitem{updp}
J.~Liu, D.~Tang, Y.~Huang, L.~Zhang, X.~Zeng, D.~Li, M.~Lu, J.~Peng, Y.~Wang, F.~Jiang \emph{et~al.}, ``Updp: A unified progressive depth pruner for cnn and vision transformer,'' \emph{arXiv preprint arXiv:2401.06426}, 2024.

\bibitem{S2ViTE}
T.~Chen, Y.~Cheng, Z.~Gan, L.~Yuan, L.~Zhang, and Z.~Wang, ``Chasing sparsity in vision transformers: An end-to-end exploration,'' \emph{Advances in Neural Information Processing Systems}, vol.~34, pp. 19\,974--19\,988, 2021.

\bibitem{kong2022spvit}
Z.~Kong, P.~Dong, X.~Ma, X.~Meng, W.~Niu, M.~Sun, X.~Shen, G.~Yuan, B.~Ren, H.~Tang \emph{et~al.}, ``Spvit: Enabling faster vision transformers via latency-aware soft token pruning,'' in \emph{European conference on computer vision}.\hskip 1em plus 0.5em minus 0.4em\relax Springer, 2022, pp. 620--640.

\bibitem{xu2024gtp}
X.~Xu, S.~Wang, Y.~Chen, Y.~Zheng, Z.~Wei, and J.~Liu, ``Gtp-vit: Efficient vision transformers via graph-based token propagation,'' in \emph{Proceedings of the IEEE/CVF Winter Conference on Applications of Computer Vision}, 2024, pp. 86--95.

\bibitem{ltmp}
M.~Bonnaerens and J.~Dambre, ``Learned thresholds token merging and pruning for vision transformers,'' \emph{Transactions on Machine Learning Research}, 2023.

\bibitem{chavan2022vision}
A.~Chavan, Z.~Shen, Z.~Liu, Z.~Liu, K.-T. Cheng, and E.~P. Xing, ``Vision transformer slimming: Multi-dimension searching in continuous optimization space,'' in \emph{Proceedings of the IEEE/CVF Conference on Computer Vision and Pattern Recognition}, 2022, pp. 4931--4941.

\bibitem{liu2022adaptive}
X.~Liu, T.~Wu, and G.~Guo, ``Adaptive sparse vit: Towards learnable adaptive token pruning by fully exploiting self-attention,'' \emph{arXiv preprint arXiv:2209.13802}, 2022.

\end{thebibliography}

\vfill

\end{document}